\documentclass[sigconf,nonacm]{acmart}
\usepackage{amsmath}
\usepackage{natbib} 
\usepackage{mathtools}
\usepackage[toc,page, title, titletoc]{appendix}
\usepackage{pgfplots}
\pgfplotsset{compat=newest}
\pgfplotsset{scaled y ticks=false}
\usepgfplotslibrary{groupplots}
\usepgfplotslibrary{dateplot}

 \usepackage{tikz}

\AtBeginDocument{%
  \providecommand\BibTeX{{%
    \normalfont B\kern-0.5em{\scshape i\kern-0.25em b}\kern-0.8em\TeX}}}
    
\begin{document}

\title[]{Anomalous NO2 emitting ship detection with TROPOMI satellite data and machine learning}

\author[S.Kurchaba]{Solomiia Kurchaba}
\email{s.kurchaba@liacs.leidenuniv.nl}
\orcid{0000-0002-0202-1898}
\affiliation{
\institution{Leiden Institute of Advanced Computer Science (LIACS)}
\institution{Leiden University}
\city{Leiden}
\country{The Netherlands}
}

\author[J. van Vliet]{Jasper van Vliet}
\email{jasper.van.vliet@ilent.nl}
\affiliation{
\institution{Human Environment and Transport Inspectorate (ILT)}
\city{Utrecht}
\country{The Netherlands}
}

\author[F.J. Verbeek]{Fons J. Verbeek}
\email{f.j.verbeek@liacs.leidenuniv.nl}
\affiliation{
\institution{Leiden Institute of Advanced Computer Science (LIACS)}
\institution{Leiden University}
\city{Leiden}
\country{The Netherlands}}

\author[C.J. Veenman]{Cor J. Veenman}
\email{c.j.veenman@liacs.leidenuniv.nl}
\affiliation{
\institution{Leiden University}
\city{Leiden}
\country{The Netherlands}}
\affiliation{
\institution{Data Science Department, TNO}
\city{The Hague}
\country{The Netherlands}
}

\date{February 2023}

\begin{abstract}
 
Starting from 2021, more demanding $\text{NO}_\text{x}$ emission restrictions were introduced for ships operating in the North and Baltic Sea waters. Since all methods currently used for ship compliance monitoring are financially and time demanding, it is important to prioritize the inspection of ships that have high chances of being non-compliant. The current state-of-the-art approach for a large-scale ship $\text{NO}_\text{2}$ estimation is a supervised machine learning-based segmentation of ship plumes on TROPOMI/S5P images. However, challenging data annotation and insufficiently complex ship emission proxy used for the validation  limit the applicability of the model for ship compliance monitoring. In this study, we present a method for the automated selection of potentially non-compliant ships using a combination of machine learning models on TROPOMI satellite data. It is based on a proposed regression model predicting the amount of $\text{NO}_\text{2}$ that is expected to be produced by a ship with certain properties operating in the given atmospheric conditions. The model does not require manual labeling and is validated with TROPOMI data directly. The differences between the predicted and actual amount of produced $\text{NO}_\text{2}$ are integrated over observations of the ship in time and are used as a measure of the inspection worthiness of a ship. To assure the robustness of the results, we compare the obtained results with the results of the previously developed segmentation-based method. Ships that are also highly deviating in accordance with the segmentation method require further attention. If no other explanations can be found by checking the TROPOMI data, the respective ships are advised to be the candidates for inspection.

\end{abstract}
\maketitle
\section{Introduction}
The industry of international shipping is one of the strongest sources of anthropogenic emission of $\text{NO}_\text{x}$ - a substance harmful both to ecology and human health. The contribution of the shipping industry to the global emission of $\text{NO}_\text{x}$  is estimated to vary between 
$15\% - 35\%$ 
\cite{crippa2018gridded, johansson2017global}, causing approximately 60,000 premature deaths annually \cite{corbett2007mortality}. To mitigate the negative impact of this industry, the International Maritime Organization (IMO) step-wisely tightens the restrictions put on emission factors of marine engines. The latest step is an 80$\%$ reduction of $\text{NO}_\text{x}$ emission for ships operating in the North and Baltic Sea \cite{AnnexVI}.

The monitoring of the compliance of ships with the IMO regulations is being performed by manual onboard inspections. However, due to the high costs, a selection of ships that will undergo inspection is needed. Among the sources of information currently used for the selection of ships are in-situ emission measurement stations \cite{beecken2014airborne, mclaren2012survey, acp-15-10087-2015} usually located at the entrance of the harbors, or airborne platform-based measurements such as planes, drones or helicopters \cite{van2016results}. The data collected with such methods give limited information on how much the selected ships emit outside of a port and are usually done near-shore. In addition, the above-mentioned methods are spot checks that usually only happen once. This does not give a possibility of having a wider perspective on ship performance.
As a result, the decisions regarding the worthiness of a ship inspection do not have sufficient justification.

The TROPOspheric Monitoring Instrument onboard the Sentinel 5 Precursor (TROPOMI / S5P) satellite launched in 2018 \cite{veefkind2012tropomi} is the first remote sensing instrument that is able to distinguish $\text{NO}_\text{2}$ plumes from individual ships \cite{georgoulias2020detection, kurchaba2021improving}. This technical improvement allows to consider remote sensing as a potential solution for ship compliance monitoring \cite{scipper}. In particular, the data from  the TROPOMI instrument could be used for the development of a data-driven inspection recommendation.

The current state-of-the-art of large-scale methods for $\text{NO}_\text{2}$ ship plume modeling use thresholding or supervised machine learning-based segmentation of TROPOMI images to attribute the measured $\text{NO}_\text{2}$ to individual ships \cite{kurchaba2021improving, rs14225809}. The latter methodology is an automated procedure improving significantly upon previously-used manual methods. However, due to the low signal-to-noise ratio of TROPOMI measurements, ship plumes are often hard to delineate, which makes the process of manual data annotation time-consuming and potentially erroneous. The absence of ground truth for a given task requires an alternative measure of validation. The ship emission proxy is an option, though often it does not cover the full list of factors that can potentially influence the levels of ship emissions. This does not allow a proper quantification of the effects of the errors coming from manual labeling. Consequently, the possibilities of the application of this approach to the task of monitoring $\text{NO}_\text{2}$ emissions from individual ships are limited.

In this study, we propose a robust method for automated selection of anomalously $\text{NO}_\text{2}$ emitting seagoing ships. 
The presented approach does not require data labeling and is validated using TROPOMI data directly. Moreover, our method is based on the integration of multiple observations, which gives a more complete perspective on ship performance.
This is achieved by training a specifically designed regression model, which predicts the amount of $\text{NO}_\text{2}$ that is expected to be registered by the TROPOMI sensor for a  given ship operating in certain atmospheric conditions. 
The difference between the predicted and actual amount of registered $\text{NO}_\text{2}$ is integrated over the available number of ship observations. The integrated difference we consider a measure of inspection worthiness of the ship.

We train the regression model with an automatically delineated Region of Interest (RoI) based on ship, wind speed, and direction.
We apply Automated Machine Learning (AutoML) to optimize the machine learning-based regression pipeline for the $\text{NO}_\text{2}$ prediction. 
To assure the robustness of the proposed method, we compare the results obtained with the regression model with the previously developed \cite{rs14225809} method for ship plume segmentation.
Ships that are also ranked as highly deviating in accordance with the ship plume segmentation model are nominated as anomalous emitters and require further attention.
We manually check the TROPOMI data for objective explanations of anomalous results. If no other explanations are found, the ships are advised to be the candidates for further inspection.

The rest of this paper is organized as follows: In Section \ref{related_work}, we present an overview of the relevant literature. In Section \ref{data}, we describe the data sources used in this study. In Section \ref{method}, we introduce the developed methodology, which is followed by the results presented in Section \ref{results}. In Sections \ref{discussion} and \ref{conclusions}, the reader can find the discussion and final conclusions respectively.

\section {Related Work}\label{related_work}
Remote sensing is a well-established technique for the measurement of emission levels. In particular, there is an extensive list of studies using satellites-predecessors of the TROPOMI / S5P to quantify the aggregated $\text{NO}_\text{2}$ emission levels produced by the shipping industry. For instance, with the  Global Ozone Monitoring Experiment (GOME) \cite{burrows1999global} satellite image, the $\text{NO}_\text{2}$ emission levels were quantified for the shipping lane between Indonesia and Sri Lanka \cite{beirle2004estimate}. The ship traces in the Red Sea were estimated using the SCanning Imaging Absorption spectroMeter for Atmospheric CartograpHY (SCIAMACHY) \cite{bovensmann1999sciamachy} onboard the  ENVIronmental SATellite (Envisat) mission \cite{richter2004satellite}. With the data from the Ozone Monitoring Instrument (OMI) \cite{levelt2006science} aboard the NASA Aura, a ship $\text{NO}_\text{x}$ emission inventory for the Baltic Sea was visualized \cite{vinken2014constraints}. Because of the low spatial resolution of the above-mentioned instruments (GOME: $40\times320$ km$^2$, SCIAMACHY: $30\times60$ km$^2$, OMI: $13\times25$ km$^2$), all the described studies were based on multi-month averaging of satellite data. While being able to identify the shipping lanes as areas with high concentrations of pollutants, these approaches are unsuitable for the quantification of emissions from individual ships. 

In \cite{georgoulias2020detection}, the authors for the first time reported that with TROPOMI satellite data, the $\text{NO}_\text{2}$ plumes produced by individual ships can be distinguished. 
The study was focused on the analysis of the plumes from the biggest ships in the area, using a basic approach for ship plume allocation. 
In \cite{kurchaba2021improving}, we showed that the application of data pre-processing techniques allows for the distinction of a greater amount of ships.
Moreover, the introduced wind error-robust method of ship-plume allocation allowed for a more accurate estimation of $\text{NO}_\text{2}$ produced by ships. 
The plume-background separation, however, was made using a univariate method of the locally optimized detection threshold. 
This approach had problems with the differentiation of plumes produced by the studied ships from all the other $\text{NO}_\text{2}$ concentration peaks within the ship's proximity.
In \cite{finch2022automated}, using elements of automatic machine learning, the authors developed a model for the automatic distinction of TROPOMI images into those that contain $\text{NO}_\text{2}$ plumes and those that do not. 
The possibilities of quantification of the intensity of the detected plumes, however, have not been provided.
In \cite{rs14225809}, we introduced a multivariate supervised learning method for automated segmentation of $\text{NO}_\text{2}$ ship plumes.
This was a notable improvement over existing baselines.
Nevertheless, the method requires manual annotation of the images. Due to the low signal-to-noise ratio of TROPOMI data, the distinguishability of the boundaries of the plume is often challenging, which makes such an annotation time-consuming and possibly misleading.
Because of the lack of ground truth, the validation of the proposed approach was performed using a theoretical ship emission proxy \cite{georgoulias2020detection} that does not take into account many factors (such as the amount of cargo on board, or local meteorological conditions) that can influence the $\text{NO}_\text{2}$ levels produced by a given ship.
As a result, validation of human labeling and consequently the approach as a whole cannot be done with the proxy. This limits the possibilities of the usage of this approach for the task of ship compliance monitoring.

In this study, we propose an approach for the automated detection of anomalously $\text{NO}_\text{2}$ emitting  ships. We specifically design a regression model for ship $\text{NO}_\text{2}$ prediction, and integrate the difference between the real and predicted amount of $\text{NO}_\text{2}$, leading to a ship emission profile. The method does not require manual labeling and is validated using TROPOMI measurements directly. To assure the robustness of the obtained results, we combine the newly developed approach with the methodology proposed in \cite{rs14225809}. 

\section{Data}\label{data}

To prepare the dataset used in this study, we combine several sources of data. We use the TROPOMI file\footnote{Open access under \url{https://s5phub.copernicus.eu/}} to retrieve an $\text{NO}_\text{2}$ tropospheric vertical column density (VCD$_{trop}$) variable, which is the objective of this study; wind data that is used to define the RoI of a ship, and as a feature of a segmentation and regression machine learning models; albedo data, as well as two VCD$_{trop}$ priors (slant column density (SCD) and air mass factor (AMS)) that are used as features of the regression model. We use Automatical Identification System (AIS) data for the positions of ships in the studied area at the moment of the satellite overpass. Finally, official ship registries are used to retrieve data about the dimensions of the studied ships.
In the following section, we provide a detailed description of all used data sources.

\subsection{TROPOMI data}\label{tropomi_sect}

\begin{figure*}
    \centering
    \includegraphics[width=1.0\linewidth]{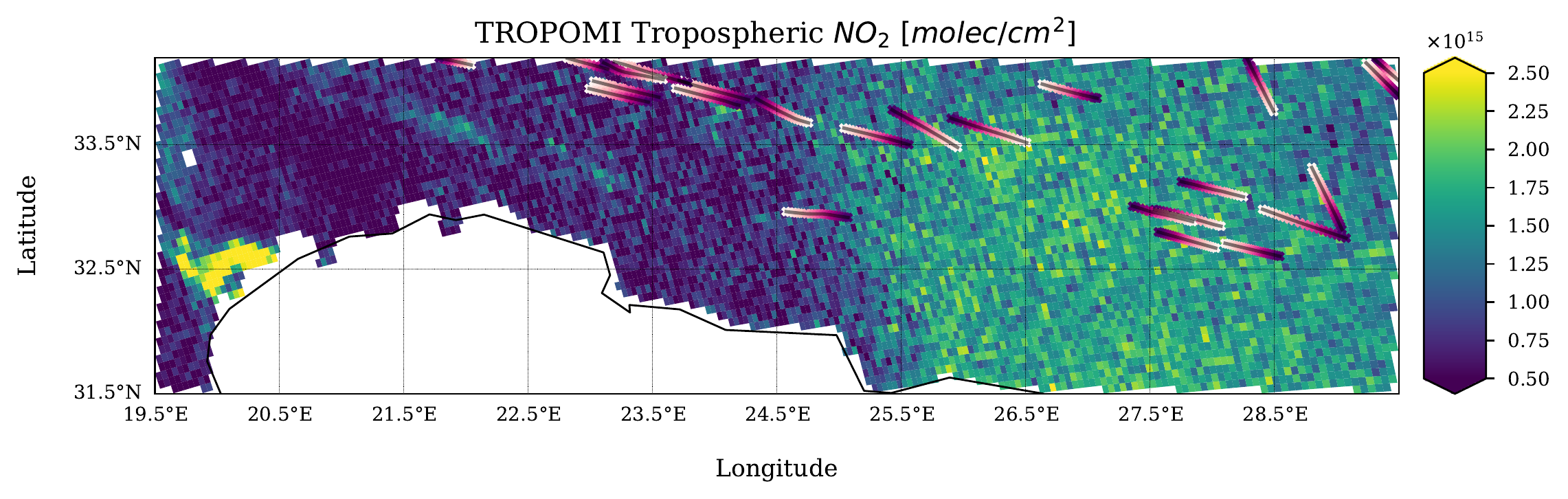}
    \caption{The $\text{NO}_\text{2}$ tropospheric column. Visualized day: June 14th, 2019. Study area: part of the Mediterranean Sea, bound by the Northern coasts of Libya and Egypt in the South and South of Crete in the North. Magenta lines indicate tracks of ships based on AIS data. The elevated background concentrations (green) in the east correspond to outflow from a variety of land-based sources. To visualize the TROPOMI data, a native local pixel size is plotted before regridding.}
\label{area_example}
\end{figure*}

TROPOMI/S5P \cite{veefkind2012tropomi} -- TROPOspheric Monitoring Instrument onboard the Sentinel 5 Precursor (S5P) satellite is a UV-Vis-NIR-SWIR (UV, visible, near-infrared, short-wave infrared) spectrometer that entered its operational phase starting from May 2018. 
It is a sun-synchronous satellite that in 24 hours performs approximately 14 orbits covering the full globe. The local equatorial overpass time of the satellite is 13:30.

The TROPOMI instrument measures spectra of multiple trace gasses, including the one of our interest -- nitrogen dioxide ($\text{NO}_\text{2}$).
The $\text{NO}_\text{2}$ gas is the most important outcome of photochemical reactions of $\text{NO}_\text{x}$ emitted by ships.
Therefore, it is suitable for compliance monitoring of ships \cite{rs14225809}. In this study, the variable of our interest is $\text{NO}_\text{2}$ tropospheric vertical column density (VCD$_{trop}$)\footnote{TROPOMI / S5P data version: 2.3.1} \cite{tropomimanual}. 

The retrieval of VCD$_{trop}$ consists of three steps \cite{tropomimanual}. The first step is spectra fitting, where the total $\text{NO}_\text{2}$ slant column density (SCD) is obtained from radiance and irradiance spectra
measured by TROPOMI instrument. During the second step, the total SCD is separated into stratospheric (SCD$_{strat}$) and tropospheric (SCD$_{trop}$) components.  Finally, as a third step, the tropospheric air mass factor (AMF) is calculated to convert the SCD$_{trop}$ into VCD$_{trop}$. The AMF is obtained on the basis of historical emission inventories (highly dependent on i.a. satellite geometry, cloud fraction, surface albedo) and a TM5-MP chemistry transport model that calculates vertical distributions of $\text{NO}_\text{2}$ at a resolution of $1\times1^{\circ}$ \cite{williams2017high}. As a result, the third step is considered to be the biggest source of the uncertainties related to VCD$_{trop}$ retrieval \cite{amt-10-759-2017,cheng2019no2}.  Among the implication of such possible uncertainties is the fact that areas with historically higher emission levels (e.g. shipping lanes or close-to-land sea regions) will tend to be more sensitive to measuring small and temporally emission sources such as plumes from individual ships. To minimize the negative impact of potential uncertainties on the obtained results, such variables as background $\text{NO}_\text{2}$ SCD, AMF, surface albedo, and sun/satellite geometry will be used as model features for ship $\text{NO}_\text{2}$ estimation.     

 The studied region is the area in the eastern Mediterranean Sea restricted by the Northern coast of Libya and Egypt from the south and South coast of Crete from the north (long: [19.5$^{\circ}$; 29.5$^{\circ}$], lat: [31.5$^{\circ}$; 34.2$^{\circ}$]). An outline of the studied area is presented in Figure \ref{area_example}. The reason for selecting this particular region for analysis is the high frequency of sunny days in the area and relatively low levels of the background $\text{NO}_\text{2}$ concentrations. Those are favorable conditions for testing the suitability of the proposed solutions for ship  $\text{NO}_\text{2}$ estimations. The study period is 20 months, starting from 1 April 2019 until 31 December 2020.

The maximal ground pixel resolution of the TROPOMI spectrometer equals $3.5\times5.5$  $\text{km}^2$ at nadir. Because of the projection of the satellite images, the real size of a pixel will vary depending on the true distance between the satellite and the captured part of the earth's surface. In order to obtain the images of a regular size, we perform regridding\footnote{The regridding is performed using the Python package HARP v.1.13.} of the original TROPOMI data into a grid of a regular size $0.045^{\circ}\times0.045^{\circ}$. For the studied area the size of the pixel after regridding translates to approximately $4.2\times5$  $\text{km}^2$.

\begin{figure}
    \centering
    \includegraphics[width=1.0\linewidth]{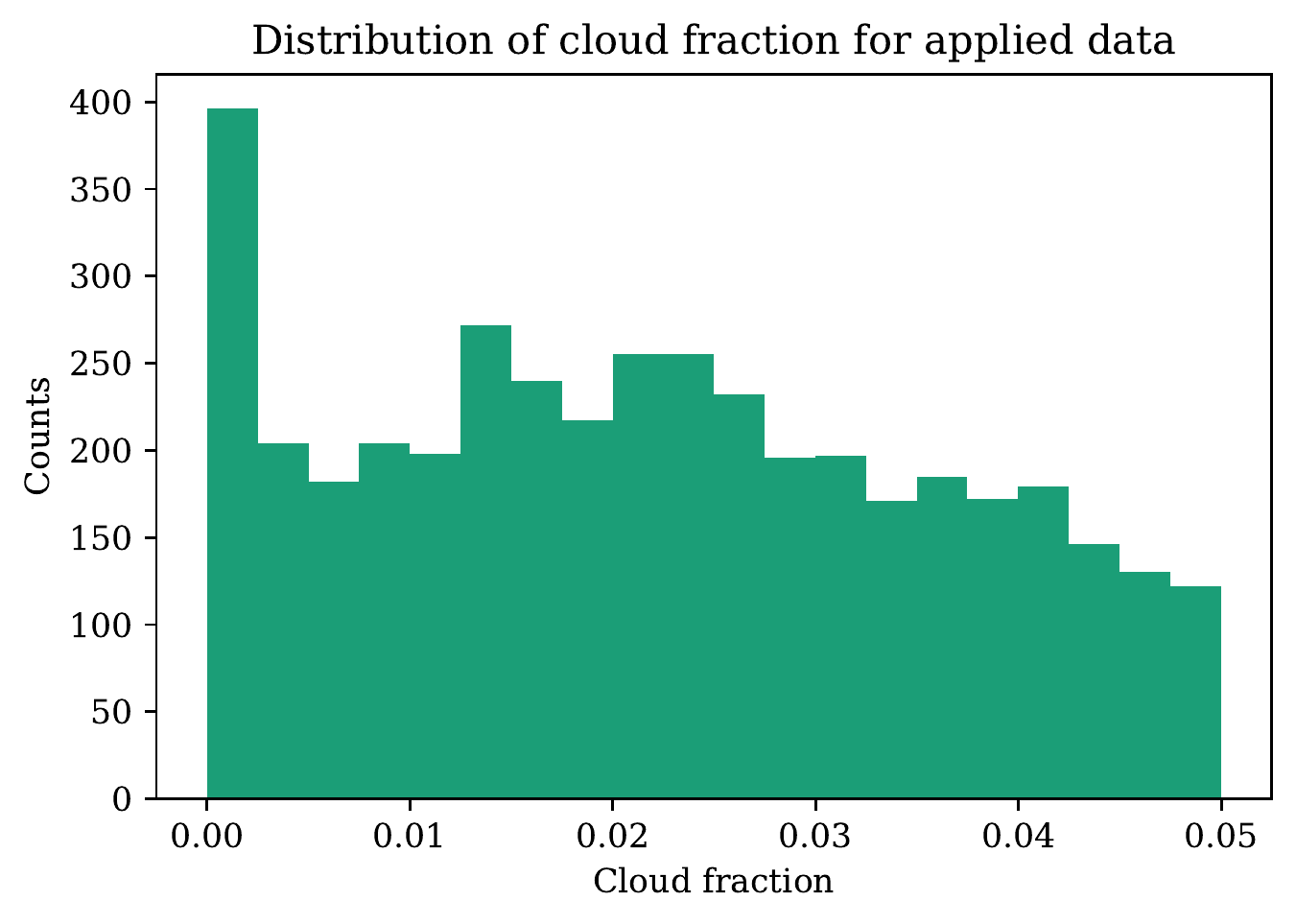}
    \caption{Histogram of the variable \textit{cloud fraction} for the dataset used in this study. Values above 0.05 were filtered out.}
\label{cloud_fraction}
\end{figure}

To assure the best possible quality of the TROPOMI retrieval, only measurements flagged with $qa\_ value> 0.5$ \cite{tropomivar}  are taken into consideration. In addition, since the TROPOMI measurements of scenes covered with clouds should not be considered valid, we filtered out from the data pixels with a cloud fraction higher than 0.05. The resulting distribution of the corresponding variable is provided in Figure \ref{cloud_fraction}. 

\subsection{Meteorological data}\label{metheo_data}

For the study, the wind information is taken from  wind speed data from the European Center for Medium-range Weather Forecast (ECMWF) at 10 m height, available with $0.25^{\circ}$ resolution at a 6-hourly time step. The used surface albedo data is the OMI minimum Lambertian equivalent reflectance (LER) at a resolution of $0.5^{\circ}$. Both ECMWF wind data and OMI surface albedo data are available as support products in the TROPOMI/S5P data file.

\subsection{Ship-related data}

Another data source used in this study is relayed through Automatic Identification System (AIS) transponders\footnote{Since 2002 all commercial sea-going vessels are obliged to carry on board an AIS transponder \cite{mou2010study}.}. The data include the position, speed, heading, and unique identifier (MMSI) of each ship carrying an active transponder.
Due to the fact that at the moment there is no open-access AIS data available, for the scope of this study, the AIS data as well as information about the dimensions of the ships (such as length, type, and gross tonnage) were provided by the Netherlands Human Environment and Transport Inspectorate (ILT). This is the Dutch national designated authority for shipping inspections, has access to commercial databases for the AIS data set used in this study, and is participating in this research.

In order to prevent the occurrence in our dataset of ships below the detection limit, we focus our analysis on the seagoing ships that are longer than 150 meters and faster than 12 kt.
Another situation we want to prevent is when too many ships contribute to the creation of the detected $\text{NO}_\text{2}$ plume, as in this case, quantification of individual contributions is extremely challenging. Thus, we remove the ships, whose trajectories within 2 hours before the satellite overpass, intersect with more than 3 other neighboring ships. This is a trade-off between a sufficient size of the dataset and the complexity of the problem of the quantification of individual contributions. 
Among all ship types present in the dataset, for the detection of anomalously emitting ships, we focus our attention on two ship types only: containers and tankers. Other ship types have not been represented in the dataset in a sufficient amount to obtain statistically significant results.

 \section{Method} \label{method}

 \begin{figure*}
       \centering
    \includegraphics[width=0.6\linewidth]{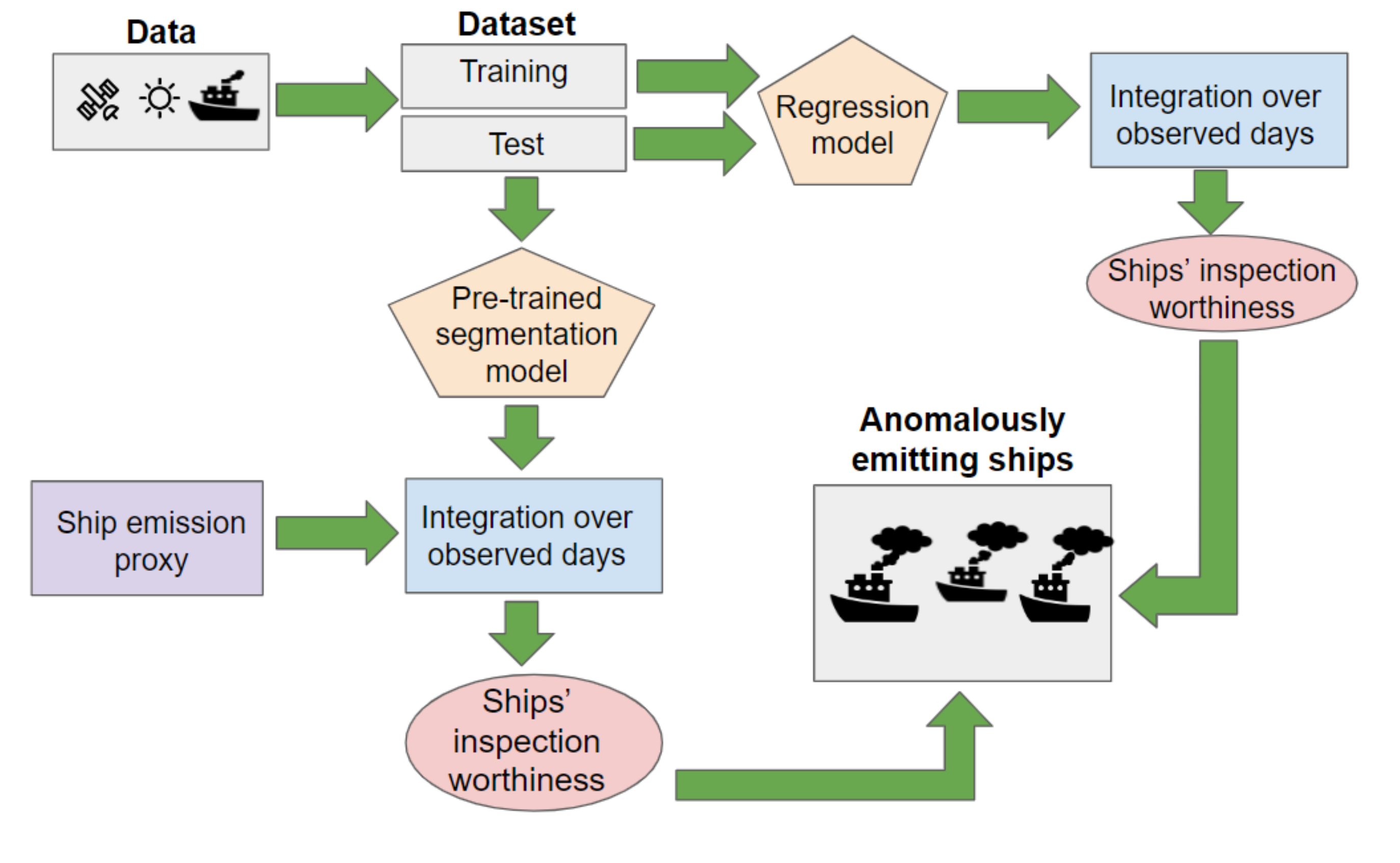}
    \caption{High-level diagram of the proposed methodology.}
\label{diagram}
  
 \end{figure*}

In this Section, we present the method for automated detection of ships that produce anomalously high amounts of $\text{NO}_\text{2}$. The method is composed of the following steps: we train a regression model for the prediction of the amount of $\text{NO}_\text{2}$ within the Region of Interest (RoI) of the analyzed ship. We calculate the difference between the observed and predicted amount of  $\text{NO}_\text{2}$ and integrate this value over all observations of the same ship within the studied period. The integrated difference between the real and predicted value of $\text{NO}_\text{2}$ we consider as a measure of the inspection worthiness of the ship. We rank the studied ships accordingly. To assure the robustness of the results, we apply the ship plume segmentation model \cite{rs14225809} to the same dataset. We compare the results obtained using the segmentation model with the value of the theoretical ship emission proxy. The results of the comparison we consider to be a measure of the inspection worthiness according to the segmentation model. The ships that are high on the inspection worthiness list of both independently trained and validated machine learning models are considered to be potentially anomalously emitting. We validate the obtained results by visual inspection of the corresponding TROPOMI measurements. Figure \ref{diagram}  provides a high-level explanation of the proposed method for the detection of anomalously emitting ships.  Below, each step of the methodology is described in detail.

\subsection{Regression model}\label{Regression_method}
Here, we describe our proposed regression model as part of a method for the detection of anomalously emitting ships. We first provide a formal definition of the proposed way for ship  $\text{NO}_\text{2}$ estimation with the regression model. Then we discuss the process of definition of an RoI of a studied ship. Finally, we introduce the details of training, optimization, and evaluation of the machine learning methodology proposed in this study.

\subsubsection{Formalization of the problem}
For a given ship $s \in S$ on a given day $d\in D$, the real amount of $\text{NO}_\text{2}$ registered by TROPOMI is calculated as:
\begin{equation}
   NO_{2;d,s} = \sum_{i\in RoI_{d, s}} VCD_{NO_{2;i}}
\end{equation}

where VCD$_{NO_2}$ is the value of the retrieved TROPOMI pixel within the RoI of the analyzed ship (see Section \ref{RoI} for more details of RoI definition).

We then use a machine learning model $f$ that based on values of features $X \in \mathbb{R}$ predicts the expected amount of $\text{NO}_\text{2}$: $\hat{NO}_{2;d,s}\in \mathbb{R}$.

\begin{equation}
    \hat{NO}_{2;d,s} = f(X_{d,s})
\end{equation}

The list of  features $X$ can be found in Table \ref{tab: feature_list}. In Appendix \ref{Aregr_dataset}, we provide histograms of distributions of the features, as well as other dataset details. 

Subsequently, we calculate a percentage difference between the predicted and real registered amount of $\text{NO}_\text{2}$:

\begin{equation}
     diff_{d,s} = 2 \cdot  \frac{NO_{2;d,s} -  \hat{NO}_{2;d,s}}{NO_{2;d,2} + \hat{NO}_{2;d,s}} \cdot 100 \%
\end{equation}

\begin{figure}
    \centering
    \includegraphics[width=1.0\linewidth]{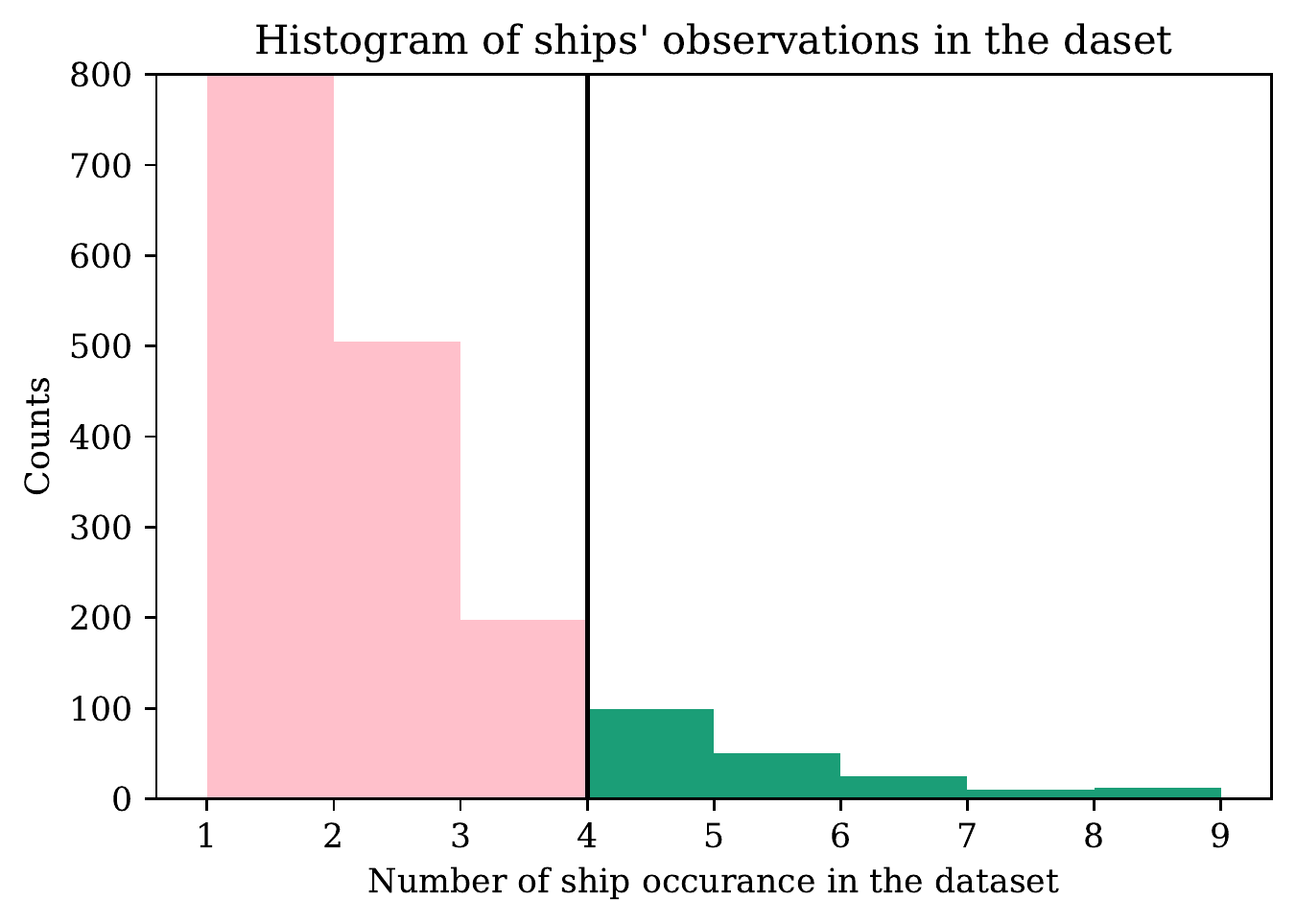}
    \caption{Histogram of occurrences of the same ship in the created dataset.  The black line indicates the set level of $min\_obs\_nb$. Only ships that have been observed  more than $min\_obs\_nb = 3$ days are taken into account for the detection of anomalously emitting ships.}
\label{nb_occur}
\end{figure}

Finally, for each ship $s \in S: |D_{s}| > min\_obs\_nb$, we integrate the obtained differences over the observed number of days  calculating arithmetic mean $\mu(diff_{s})$ and standard deviation $\sigma(diff_s)$ of the observed differences.
The $|D_{s}|$ is the number of days when the ship $s$ was observed, $min\_obs\_nb$ is the minimal number of days we require the ship to be present in the dataset so that its profile is representative enough to make the decision about being anomalously emitting. Figure \ref{nb_occur} represents the number of ship occurrences in the dataset used for the training of the regression model. In this study\footnote{Providing more ship observation data is available, the level of $min\_obs\_nb$ can be increased further.}, we set $min\_obs\_nb = 3$. 

A high value of $\mu(diff_{s})$ represents a situation when the real value of $\text{NO}_\text{2}$ for a given ship was repeatedly underestimated by the model. This means that the amount of $\text{NO}_\text{2}$ that was registered by the TROPOMI for a given ship was consistently higher than can be expected given the ship's characteristics and operational atmospheric conditions. Thus, we consider $\mu(diff_{s})$ to be a measure of the inspection worthiness of the ship in accordance with the regression model $IW^{regr}_{s}$:
\begin{equation}
IW^{regr}_{s} = \mu(diff_{s})
\end{equation}

The value $\sigma(diff_s)$ is a measure of the consistency of the obtained results. 
Since the satellite measurement results have a lower limit and do not have an upper limit, a very high $\sigma(diff_s)$ can only occur from the fact that very high values of $\text{NO}_\text{2}$ were assigned to a ship that on a regular basis does not produce that much -- only high $\text{NO}_\text{2}$ outliers can cause a high standard deviation.
Such a situation is not of our interest. Therefore, ships with outlying values of $\sigma(diff_s)$ will be removed from the analysis. The value of $\sigma(diff_s)$ is considered to be outlying if $\sigma(diff_s) > \mu(\sigma(diff_s)) + 2\sigma(\sigma(diff_s))$, which corresponds to $5\%$ of the highest observations of $\sigma(diff_s)$.

\begin{table}[]
    \centering
    \begin{tabular}{cc}
    \toprule
         \textbf{Feature type} & \textbf{Feature name}\\ 
         \midrule
          Ship related & Ship length\\ 
          & Ship speed\\ 
          & Ship heading\\ 
          & Gross tonnage\\ 
          & Ship type\\ 
          \hline
          State of the atmosphere & Wind speed\\ 
          & Wind direction\\ 
          & Surface albedo\\
          & Solar zenith angle\\ 
          & Measurement month\\ 
          \hline
          Priors for background & Average $\text{NO}_\text{2}$ VCD$_{trop}$ outside RoI \\
          & Average $\text{NO}_\text{2}$ SCD outside RoI \\
          & AMF outside RoI\\ 
          & Sensor zenith angle\\ 
          
    \bottomrule
    \end{tabular}
    \caption{List of features used for the regression model.}
    \label{tab: feature_list}
\end{table}

\subsubsection{Defining region of interest}\label{RoI}

\begin{figure*}
    \includegraphics[width=0.9\linewidth]{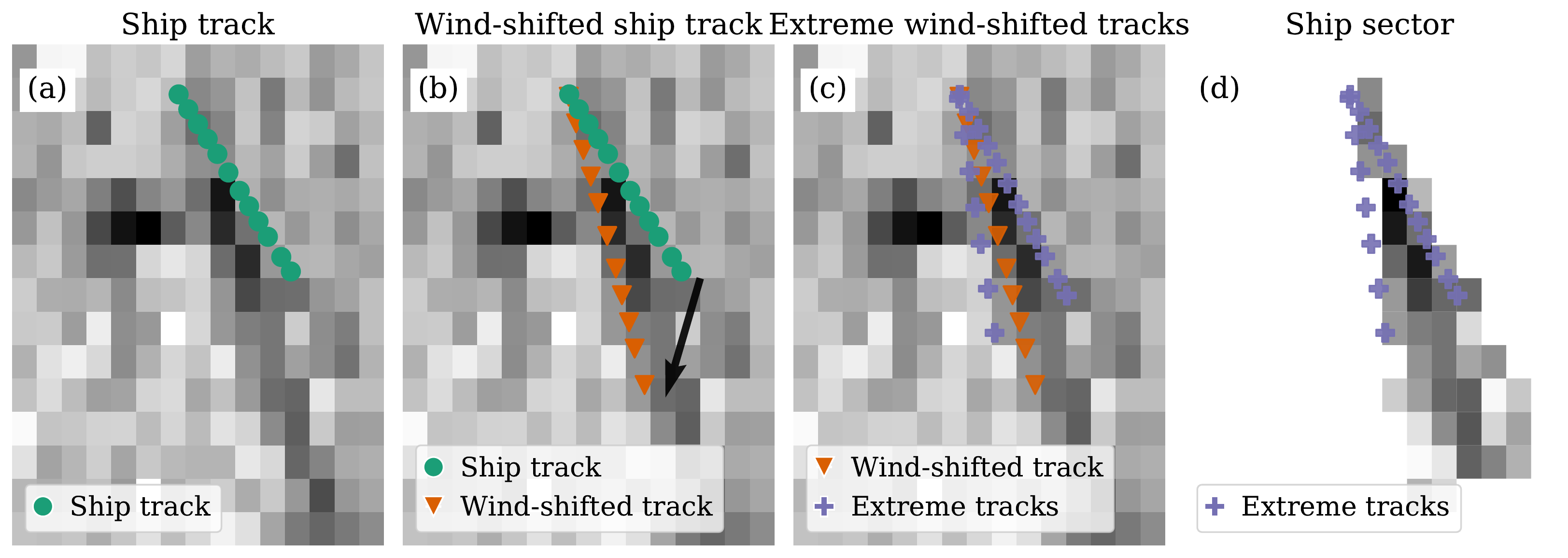}
    \caption{Ship sector definition pipeline. 
    Background -- the TROPOMI $\text{NO}_\text{2}$ signal for the area around the analyzed ship. Two ship plumes can be distinguished, but only one is of interest. (a) \textit{Ship track} -- estimated, based on AIS data records. The ship track is shown for the time period starting from 2 hours before until the moment of the satellite overpass. (b) \textit{Wind-shifted ship track} --  a ship track shifted in accordance with the speed and direction of the wind. It indicates the expected position of the ship plume. A black arrow indicates the wind direction. (c) \textit{Extreme wind-shifted ship tracks} -- calculated, based on wind information with assumed uncertainties; define the borders of the \textit{ship sector}. (d) A resulting \textit{ship sector} -- an ROI of an analyzed ship. For all presented images, the size of the pixel is equal to 4.2 $\times$ 5 km$^2$.}
\label{image_pipeline}
\end{figure*}

The RoI of the ship defines a region within which the concentration of $\text{NO}_\text{2}$ caused by a given ship will be predicted. In this study, we use a method of RoI assignation presented in \cite{kurchaba2021improving}. 

First, we estimate the trajectory of the ship -- a \textit{ship track} -- using AIS ship data, starting from two hours before, until the moment of the satellite overpass (c.f. Figure \ref{image_pipeline}a). The observation duration of two hours was selected considering an average lifetime of $\text{NO}_\text{x}$ \cite{DEFOY20151}.

Secondly, we assume that the plume emitted by a ship has moved in accordance with wind direction by a distance $d = v \times \lvert \Delta t \rvert$, where $v$ is the local wind speed for a coinciding time, and $\lvert \Delta t \rvert$ is a time difference between the time of the satellite overpass and the time of a given AIS ship position. 
In this way, we obtain a trajectory that we call a \textit{wind-shifted ship track}. An illustration of a \textit{wind-shifted ship track} is depicted in Figure \ref{image_pipeline}b. 

Both wind speed and wind direction are assumed to be constant for the whole time during which we study the plume. Such an assumption may create uncertainties in the expected position of the plume of the ship. Therefore, in the third step, we calculate the \textit{extreme wind-shifted tracks}, by adding the margin of wind-related uncertainty to each side of the \textit{wind-shifted ship track} -- c.f. Figure \ref{image_pipeline}c. The \textit{extreme wind-shifted tracks} define the borders of the RoI of the analyzed ship that we refer to as a \textit{ship sector}. The \textit{ship sector} delineates the area within which we study the plume produced by the analyzed ship. In Figure \ref{image_pipeline}d  an example of a resulting RoI that we call a \textit{ship sector} is presented. 

\subsubsection{Model optimization}\label{regr_model_opt}

\begin{figure*}
    \centering
    \includegraphics[width=0.8\linewidth]{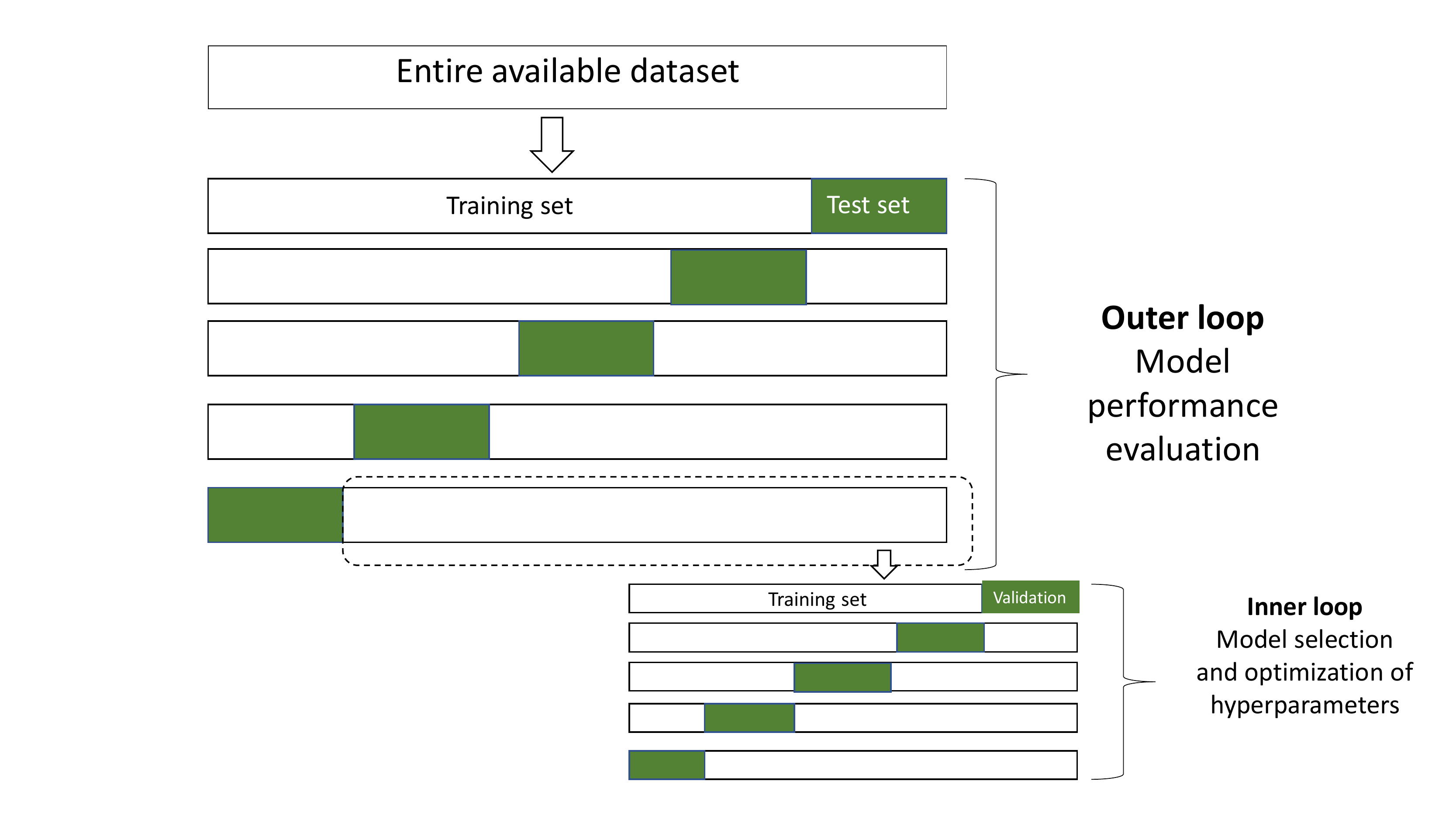}
    \caption{Applied scheme of cross-validation. In the outer loop, we generated five test sets that were used for the regression model performance evaluation, as well as for the detection of anomalously emitting ships. In the inner loop, the generated training and validation sets were used by AutoML algorithms to optimize machine learning pipelines for regression models.}
\label{cv_scheme}
\end{figure*}

In this study, we use a nested scheme of cross-validation (see Figure \ref{cv_scheme}). Within the outer 5-fold loop of cross-validation we create 5 "hold out" non-overlapping test sets and 5 training and validation sets. The test sets are used for:
\begin{enumerate}
    \item Performance evaluation of the regression model.
    \item Detection of anomalously emitting ships.
\end{enumerate}
We use the training and validation datasets for the optimization of the regression model, which is happening within the inner loop of cross-validation. 

The task of model optimization is tackled  using automated machine learning (AutoML) \cite{hutter2019automated}. With AutoML, we aim to solve a so-called CASH problem, which stands for Combined Algorithm Selection and Hyperparameter optimization \cite{kotthoff2019auto}. Given the absence of available benchmarks for our original dataset, such a technique allows for an efficient selection of a regression model and feature preprocessor from among a wide variety of machine learning models and feature transformation techniques without preliminary model selection experiments. In this study, we address the CASH problem using  TPOT (Tree-based Pipeline Optimization Tool) \cite{olson2016evaluation} -- a Python package for automatic selection of machine learning pipelines based on genetic programming (GP) \cite{koza1994genetic}. 

The results obtained using the TPOT AutoML library are benchmarked towards the results obtained using the eXtreme Gradient Boosting (XGBoost) \cite{chen2016xgboost} regression model with the default hyperparameters settings. 
The XGBoost model is considered to be a good choice when it comes to tabular data \cite{2207.08815},  as well as showed the best performance on the same type of data in our previous study \cite{rs14225809}. 
We, therefore, have adapted it as a baseline method.

\subsubsection{Model evaluation metrics}\label{eval_metrics}
In this study, we used three measures of regression model quality.
The first one is the Pearson correlation coefficient -- a measure of linear correlation between two variables. 
The Pearson correlation coefficient $\rho$ is defined as:
\begin{equation}
    \rho = \frac{Cov(Y, \hat{Y})}{\sigma(Y)\sigma(\hat{Y})}
\end{equation}
where $Cov$ is the covariance, $\sigma(Y)$, and $\sigma(\hat{Y})$ are standard deviations of real and predicted values of a target variable respectively. The value  $\rho=1$ indicates a perfect linear correlation, value $\rho=-1$ indicates perfect linear anti-correlation, and  $\rho=0$ is the total absence of linear correlation. 

The second measure used is a coefficient of determination $R^2$. The $R^2$ is defined as:
\begin{equation}
    R^2 = 1 -  \frac{\sum (y_{i} - \hat{y}_{i}) ^ 2}{\sum (y_{i} - \bar{y})^2}
\end{equation}
The $R^2 \in [0; 1]$ is a measure of the goodness of fit of a model and is interpreted as the part of the variation of the predicted variable that is explained by the regression model. The  $R^2=1$ suggests that the predictions obtained with a regression model fit the data perfectly well. In this study, we use $R^2$ as the objective function for all regression models evaluated during the AutoML selection of the optimal regression model.

The third evaluation metric is a Spearman correlation coefficient. Assuming $R(Y)$ and $R(\hat{Y})$ are ranks of the corresponding variables, we define the Spearman correlation coefficient  $r_s\in[-1;1]$ as:

\begin{equation}
    r_s = \rho(R(Y)R(\hat{Y}))
\end{equation}
The Spearman coefficient tells how well a relationship between real and predicted values of a target variable can be described by a monotonic function. A perfect Spearman correlation of $r_s=1$ or $r_s=-1$ is achieved when one variable is a perfect monotonic function of the other.

\subsection{Detection of anomalously emitting ships}

In order to assure the robustness of the proposed method for detecting anomalously emitting ships, we compare the results obtained with the regression model with another, independently trained and validated machine learning model applied to the same dataset. We intersect the results obtained with both considered models in order to obtain a list of potentially anomalously emitting ships. Hereafter, we introduce the ship plume segmentation model \cite{rs14225809} that is added to the presented regression model as a decision support tool, and explain how the results of both models are used to make a decision regarding the candidate selection of anomalously emitting ships.

\subsubsection{Segmentation Model}\label{Segmentation_method}
As a support tool for the presented regression model, we use the ship plume segmentation model prepared in accordance with the methodology introduced in  \cite{rs14225809} and trained using the dataset (for details, see Appendix \ref{Asegm_dataset}) provided in the same study. Here, we provide a formal introduction of the used approach.

For a given ship $s \in S$ on a given day $d\in D$, the estimated amount of $\text{NO}_\text{2}$ can be expressed as:

\begin{equation}
   \hat{NO}_{2;d,s} = \sum_{i \in RoI_{d,s}}\hat{y}_i \cdot \hat{NO}_{2,i},
\end{equation}
where RoI is defined in the same way as it was explained for the regression model (see Section \ref{RoI}), and $\hat{y}_i \in \{0, 1\}$ is an output of the segmentation model for a TROPOMI pixel $i$ within a RoI of the ship $s$ on a day $d$. Details on the segmentation model can be found in Appendix \ref{Asegm_model}.

Following the methodology from \cite{rs14225809}, to validate the results obtained with the segmentation model, we use a theoretical ship emission proxy $E_{d,s}$ \cite{georgoulias2020detection} defined as:

\begin{equation}
     E_{d, s} = L_s^2 \cdot u_{d, s}^3
\end{equation}
where $L_s$  is the length  of the ship $s$ in $m$, and $u_{d,s}$ is its average speed on a day $d$ in $m/s$, derivation details see \cite{georgoulias2020detection}.  

For each ship $s\in S: |D_{s}| > min\_obs\_nb$, we integrate the $\hat{NO}_{2;d,s}$ and $E_{d, s}$ over the days of observation by calculating their arithmetic means $\mu(\hat{NO}_{2;d,s})$ and $\mu(E_{d, s})$ respectively.

We assume that $\mu(\hat{NO}_{2;s})$ is linearly proportional to $\mu(E_s)$. Therefore, we can express it as:

\begin{equation}
    \mu(\hat{NO}_{2;s}) = \alpha \cdot \mu(E_s) + \beta + \epsilon_s, 
\end{equation}
where $\alpha$ and $\beta$ are the parameters of the fitted linear equation. We consider $\epsilon_s$ the measure of the inspection worthiness of the ship in accordance with the segmentation model:

\begin{equation}
    IW^{segm}_{s} = \epsilon_s
\end{equation}

In the case of the ship plume segmentation model, the measure of consistency of the results is a standard deviation of estimated values of $\text{NO}_\text{2}$, $\sigma(\hat{NO}_{2;s})$. The ships for which $\sigma(\hat{NO}_{2;s}) > \mu(\sigma(\hat{NO}_{2;s})) + 2\sigma(\sigma(\hat{NO}_{2;s}))$ are considered to be outlying and will not be taken into consideration.

\subsection{Merge of two models to identify anomalous ships}
In order to identify anomalously emitting ships, we intersect the results obtained with the two independently trained/validated machine learning models: a newly developed regression model for the prediction of ship's $\text{NO}_\text{2}$ within the assigned RoI, and ship plume segmentation model developed in previous study \cite{rs14225809}.  To assure the comparability of the results, we perform a normalization of the inspection worthiness measures obtained from both used methods, defining $norm\_IW^{regr}_{s}, norm\_IW^{segm}_{s} \in [0,1]$. The normalization is performed using min-max scaling applied on $IW_{regr_{s}}$ and  $IW_{segm_{s}}$ such that:

\begin{equation}
    norm\_IW^{regr}_{s} = \frac{IW^{regr}_{s}-min(IW^{regr}_{s})}{max(IW^{regr}_{s}) - min(IW^{regr}_{s})}
\end{equation}
\begin{equation}
norm\_IW^{segm}_{s} = \frac{IW^{segm}_{s}-min(IW^{segm}_{s})}{max(IW^{segm}_{s}) - min(IW^{segm}_{s})}
\end{equation}

Providing a decision threshold $t$, the ship is assigned to the list of anomalously emitting ships in accordance with the following rule: 
\begin{equation}
\begin{multlined}
     norm\_IW^{regr}_{s} > t  \land norm\_IW^{segm}_{s} >t \iff s \in Anomalous\_emitters,
     \end{multlined}
\end{equation}
such that:

\begin{align}
   & Anomalous\_emitters = \{s_1, ..., s_n\}: \nonumber \\ 
   & norm\_IW^{regr}_{s_i} \cdot norm\_IW^{segm}_{s_i} < norm\_IW^{regr}_{s_{i+1}} \cdot norm\_IW^{segm}_{s_{i+1}}
\end{align}

The decision about the selection of the used threshold level $t$ is left to the user. In this study, the threshold was manually selected as $t=0.55$.

\section{Results}\label{results}

In this Section, we present the obtained results. We first present the results of the regression model optimization. We then show the aggregated results of the application of the regression and segmentation models and perform the selection of potentially anomalously emitting ships. Finally, using a one-way ANOVA analysis of group differences, we inspect the obtained results on the presence of a decision bias resulting from the merge of regression and segmentation models. 

\begin{table}[]
    \centering
    \begin{tabular}{cccc}
    \toprule
    \textbf{Method} & \textbf{Pearson} & \textbf{Spearman} & \textbf{R$^2$} \\
    \midrule
         TPOT & 0.740 $\pm$ 0.058 &   0.664 $\pm$ 0.042 &  0.538 $\pm$ 0.08 \\
         Default XGBoost & 0.715 $\pm$ 0.057 &  0.622 $\pm$ 0.037 & 0.497 $\pm$ 0.098 \\
        
    \bottomrule
    \end{tabular}
    \caption{Regression model results. Hyperparameters applied for AutoML optimization: Maximal evaluation time: 10 min; Population size: 50; Number of generations: 50; Early stopping criteria: 10.}
    \label{tab:regr_results}
\end{table}

\begin{table}[]
    \centering
    \begin{tabular}{cc}
    \toprule
     \textbf{Feature processor} & \textbf{Model} \\
    \midrule
       MaxAbs Scaler & Gradient Boosting \cite{friedman2002stochastic} \\
         MaxAbs Scaler & Gradient Boosting  \\
         Polynomial Features (2$_{nd}$ deg.) & XGBoost \cite{chen2016xgboost} \\
         Standard Scaler & Gradient Boosting \\
         Standard Scaler & XGBoost \\
    \bottomrule
    \end{tabular}
    \caption{A model and a feature pre-processor selected by TPOT as optimal at a given iteration of cross-validation.}
    \label{tab:best_models}
\end{table}

\subsection{Regression model optimization} 

In Table \ref{tab:regr_results}, we present the results of the regression model optimization. The application of the TPOT pipeline optimization algorithm allowed us to substantially improve the results over the default XGBoost regression model that  was used as a baseline, in accordance with all three used quality measures. In Table \ref{tab:best_models}, we provide the models and feature pre-processing methods that were selected as optimal (best performance on validation set) at each iteration of cross-validation. The XGBoost model was still one of the most often selected optimal models. Even in this scenario, the advantage of the application of AutoML was gained by the possibility of selection of feature pre-processing method, as well as by the performance of automatic optimization of the hyperparameters of the model such as a number of estimators, maximal depth, learning rate, subsample, and minimal child weight. Another selected model was the related Gradient Boosting algorithm. 

\subsection{Detection of anomalously emitting ships}

\begin{figure}
   \centering
    \includegraphics[width=1.0\linewidth]{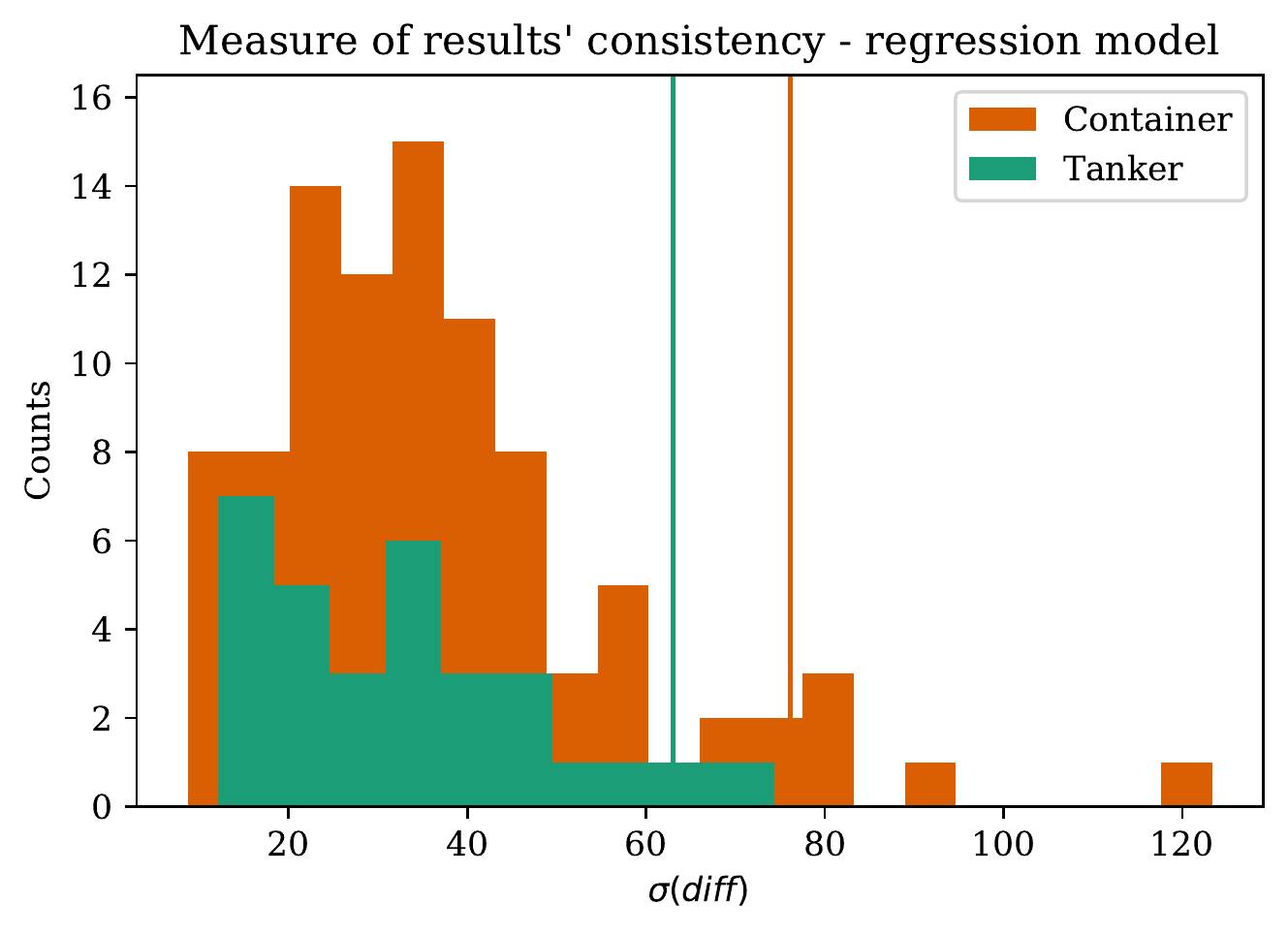}
    \caption{A measure of the consistency of results of the regression model $\sigma(diff_{s})$. The ships $s$ for which the values of $\sigma(diff)$ are higher than the threshold indicated by the vertical lines will be removed from the analysis. }
\label{regr_std_hist}
\end{figure}

\begin{figure}
   \centering
    \includegraphics[width=1.0\linewidth]{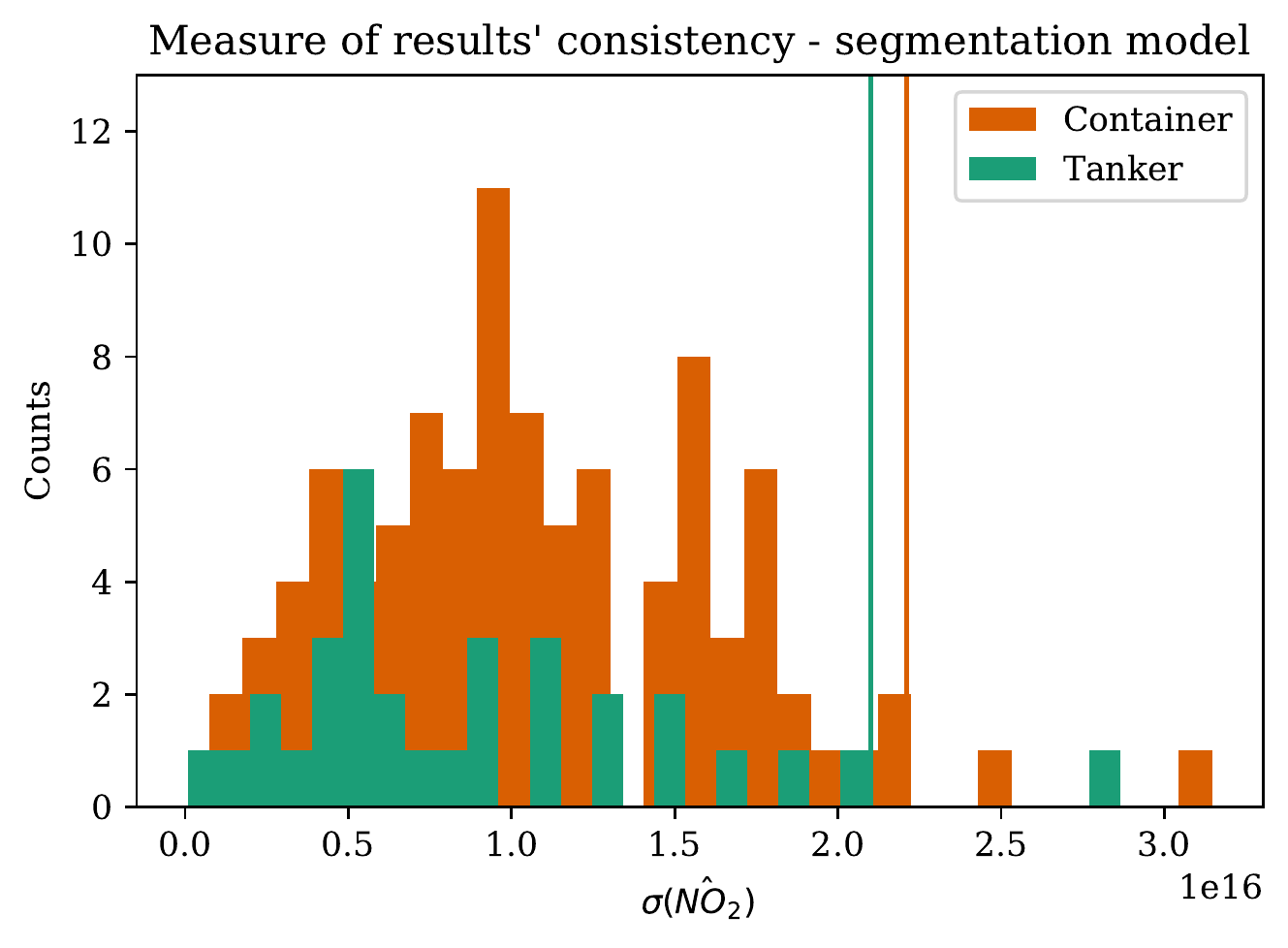}
    \caption{A measure of the consistency of results of the ship plume segmentation model $\sigma(\hat{NO}_{2;s})$. The ships $s$ for which the values of $\sigma(\hat{NO}_{2})$ are higher than the threshold indicated by the vertical lines will be removed from the analysis. }
\label{segm_std_hist}
\end{figure}

\begin{figure*}
    \centering
    \includegraphics[width=0.8\linewidth]{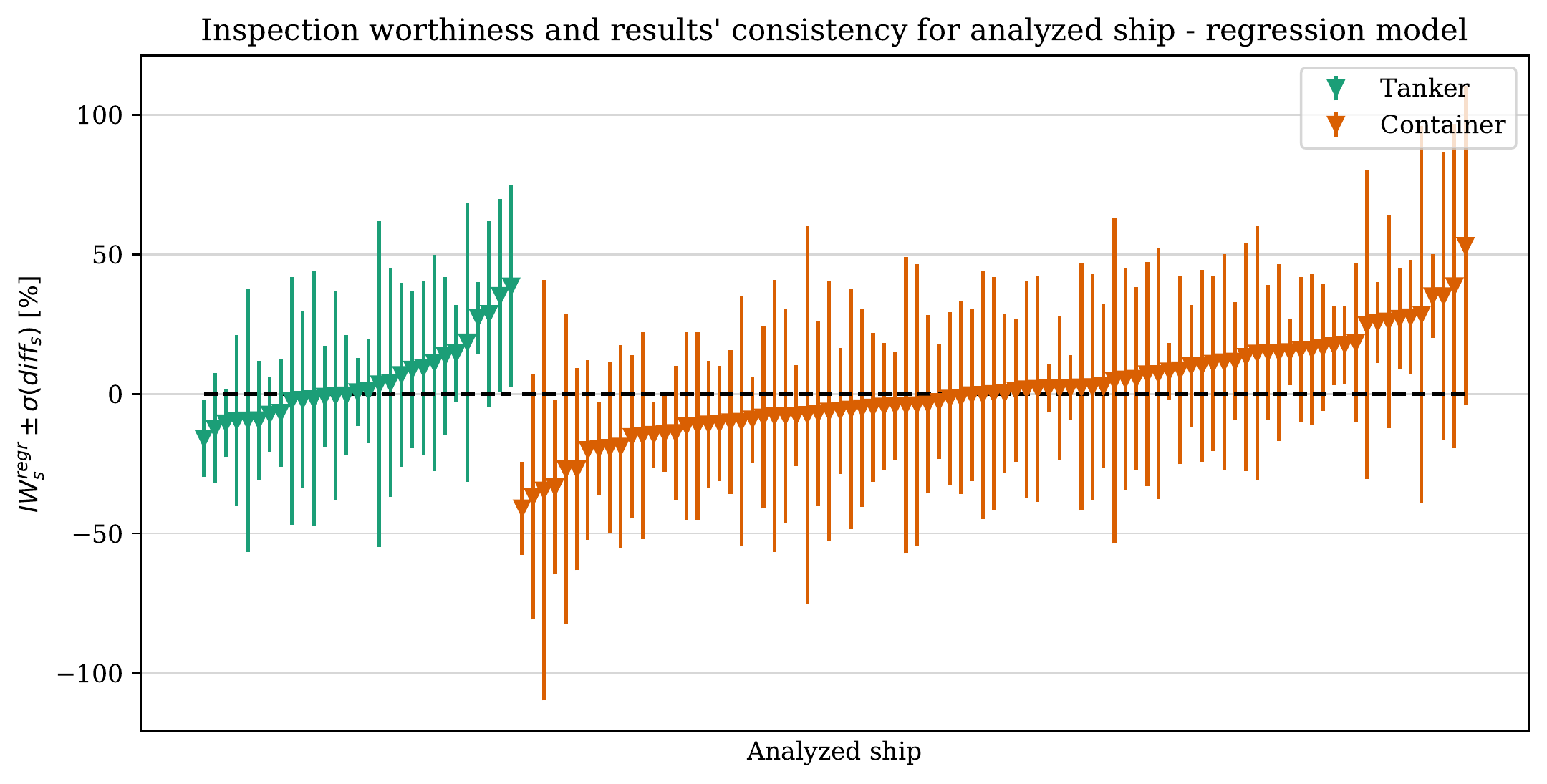}
    \caption{The triangle-shaped markers indicate the measure of ship inspection worthiness in accordance with the regression model $IW^{regr}_{s}$. The vertical lines indicate $\sigma(diff_{s})$ - the measure of the consistency of results for a given ship.}
\label{regr_diff_med}
\end{figure*}

\begin{figure*}
    \centering
    \includegraphics[width=0.8\linewidth]{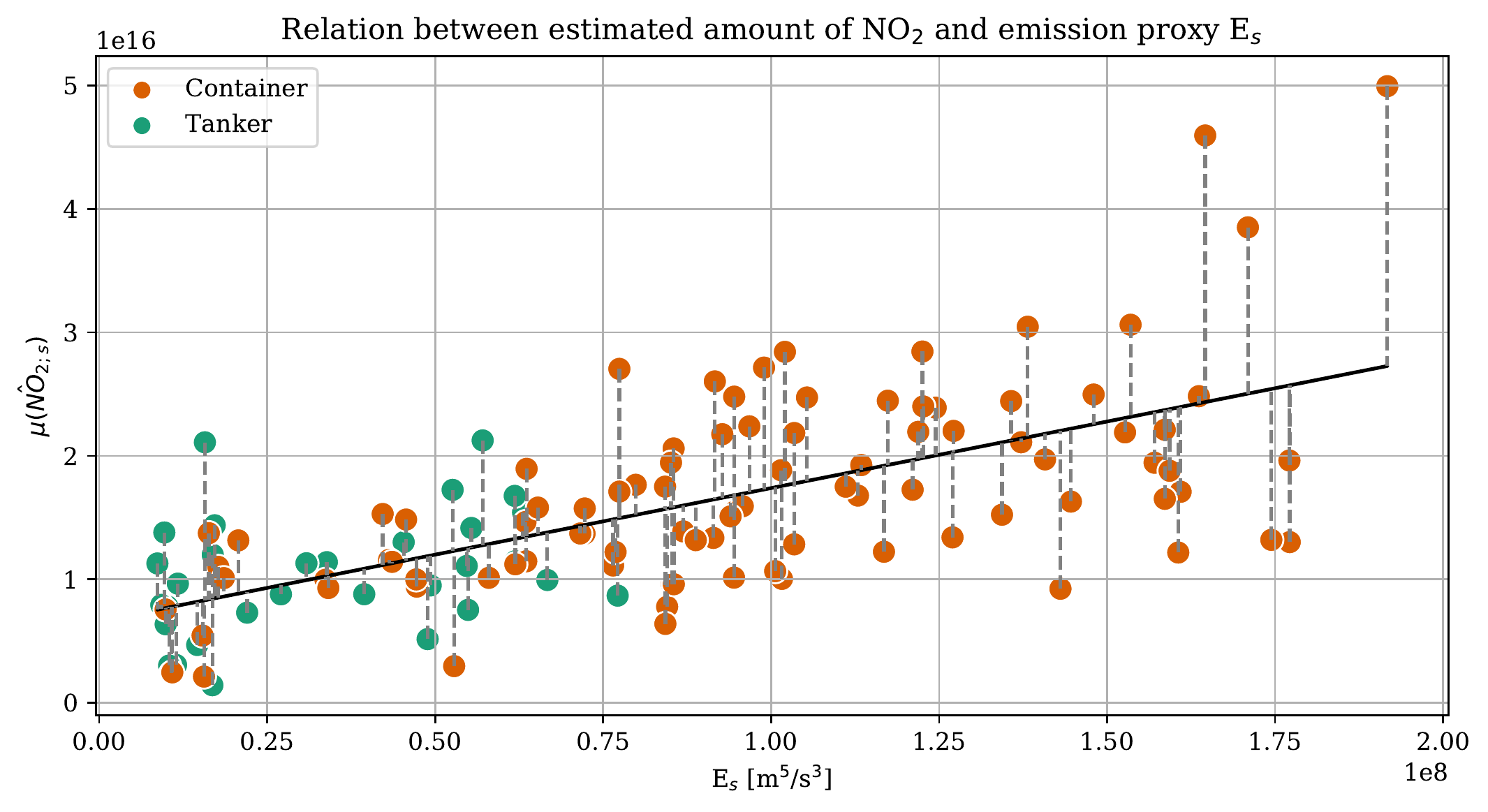}
    \caption{Relation between the estimated amount of $\text{NO}_\text{2}$ using the segmentation model and ship emission proxy with a fitted linear trend. Gray dashed lines indicate the measure ship inspection worthiness $IW^{segm}_{s}$ according to the plume segmentation model.}
\label{proxy}
\end{figure*}

Here, we analyze the results of the application of the regression and plume segmentation model with the aim of detecting anomalously emitting ships. First, for each model, we calculated the measures of the consistency of the results, i.e. $\sigma(diff_{s})$  and $\sigma(\hat{NO}_{2;s})$, removing the resulting outlying values from the analysis. The histograms of the consistency measures of the regression and the segmentation models along with the cut-off thresholds are presented in Figures \ref{regr_std_hist} and \ref{segm_std_hist} respectively.

In Figure \ref{regr_diff_med}, we depict the integrated results of the regression model for each studied ship ($\mu(diff_{s})$,  $\sigma(diff_{s})$) and rank them in ascending order of inspection worthiness, $IW^{regr}_{s} = \mu(diff_{s})$. Ships for which the difference between the real and predicted values of $\text{NO}_\text{2}$ is strongly positive are the most interesting for us. Figure \ref{proxy} presents the 
 resulting relationship between the averaged amounts of  $\mu(\hat{NO}_{2;s})$ for each ship and averaged ship emission proxy $\mu(E_s)$. The black line indicates the fitted linear trend. The gray dashed lines indicate the ship inspection worthiness $IW^{segm}_{s}$ in accordance with the segmentation model. The ships for which the $IW^{segm}_{s}$ is the highest are of our main interest.

 \begin{figure}
    \centering
    \includegraphics[width=1.0\linewidth]{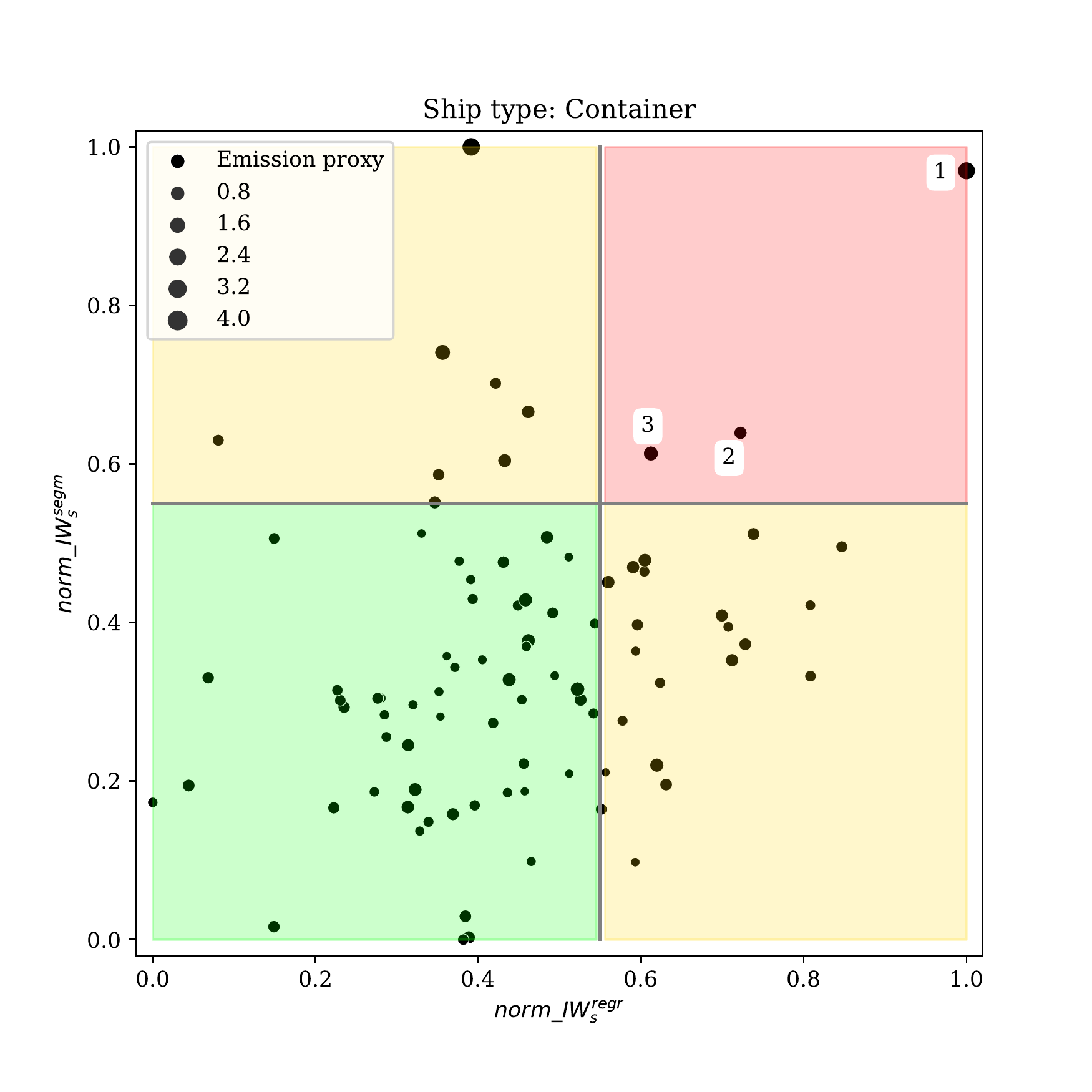}
    \caption{ Ship type: Container. Combination of results of segmentation and regression models. Values of the inspection worthiness obtained from each model were normalized using min-max scaler.}
\label{two_models_med_container}
\end{figure}

\begin{figure}
    \centering
    \includegraphics[width=1.0\linewidth]{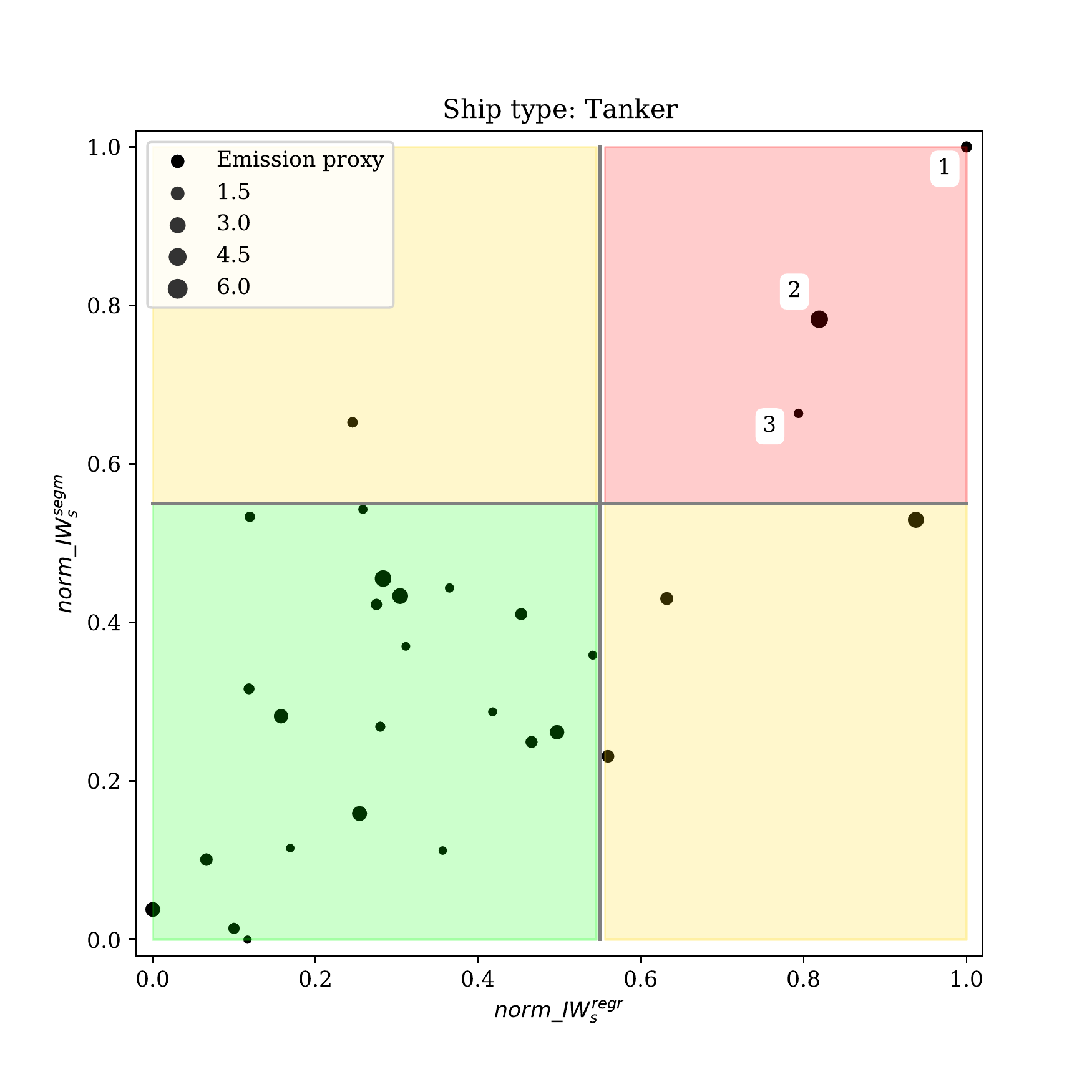}
    \caption{Ship type: Tanker. Comparison between results of segmentation and regression models. Values of the inspection worthiness obtained from each model were normalized using min-max scaler.}
\label{two_models_med_tanker}
\end{figure}

As a next step, we combine the errors obtained from the regression model with the errors of the ship plume segmentation model. Figures \ref{two_models_med_container}, \ref{two_models_med_tanker} visualize the combined inspection worthiness for containers and tankers respectively.
Each of the figures is split into four zones: green, red, and two yellow.
The size of the ship markers is scaled in accordance with the average value of the emission proxy calculated for this ship. 
The ships located in the green zone of the plots, we consider as weak emitters, because  both of the models overestimate the actual level of  $\text{NO}_\text{2}$ produced by the respective ship. 
The two yellow zones of the figures indicate ships for which one of the models overestimates the actual level of  $\text{NO}_\text{2}$, while the other model underestimates the actual level. 
This can be due to the low resistance of the particular machine learning model to certain types of difficult modeling conditions as, for instance, a combination of land-based $\text{NO}_\text{2}$ sources, plume accumulated within one TROPOMI pixel, etc. Finally, the red zone of the plot indicates ships for which the measures of inspection worthiness of both models achieve the highest values.  We call those ships potentially anomalously emitting since throughout two years of analysis they were producing more than is expected based on their characteristics and operational atmospheric conditions. 
Clearly, to make final conclusions, the detected ships should be studied closer.

\subsection{Visual verification of potential anomalous emitters}
\begin{table}[]
    \centering
    \begin{tabular}{cccc}
    \toprule
        \textbf{Ship type} & \textbf{Ship Id}  & \textbf{$\sigma(diff)$} & \textbf{$\sigma(\hat{NO}_2)$}  \\
        \midrule
        Tanker & 1 & 0.36 & 2.03 $\cdot 10^{16}$ \\
         & 2& 0.33 & 1.4 $\cdot 10^{16}$ \\
         & 3& 0.12 & 0.65  $\cdot 10^{16}$ \\
        \hline
        Container & 1 & 0.57 & 1.5 $\cdot 10^{16}$ \\
        & 2 & 0.17 & 0.99  $\cdot 10^{16}$ \\
        & 3 & 0.22 & 1.5 $\cdot 10^{16}$ \\
    \bottomrule   
    \end{tabular}
    \caption{Measures of results consistency of regression ($\sigma(diff)$) and segmentation ($\sigma(\hat{NO}_2)$) models, for ships identified as anomalous emitter. Ship Ids are in accordance to the numbers assigned in Figures \ref{two_models_med_container} and \ref{two_models_med_tanker} for containers and tankers respectively.}
    \label{tab:stds}
\end{table}

\begin{figure*}
    \centering
    \includegraphics[width=1.0\linewidth]{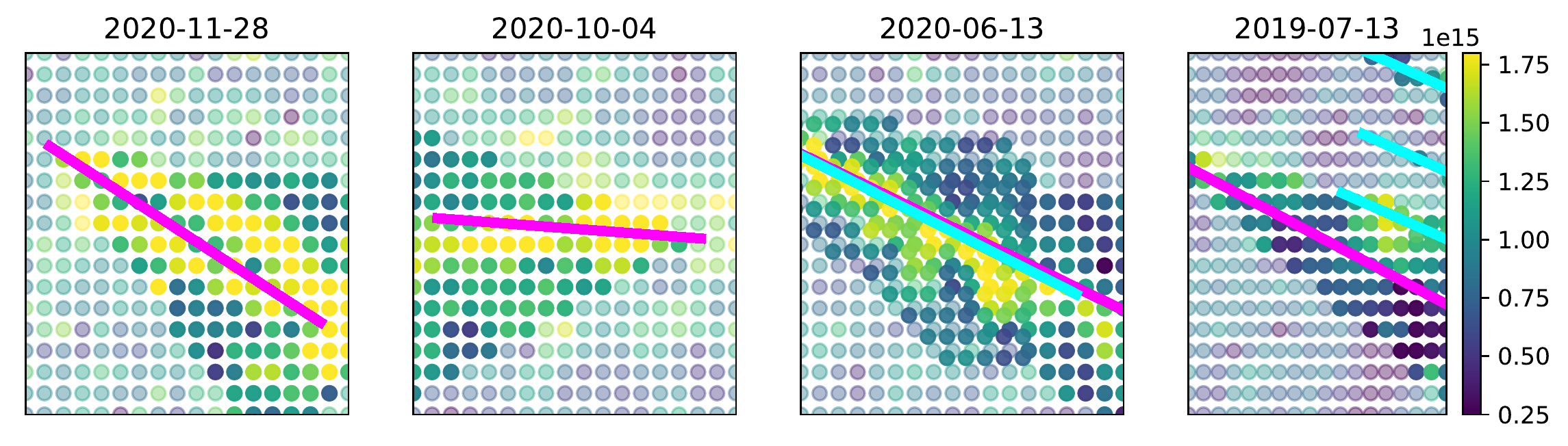}
    \caption{Ship type: Container. Example of an outlying ship 1.  Ship length: 398 m. Average ship speed: 19.6 kt. Year of built: 2008. Lines represent shifted ship tracks. Magenta line -- ship of interest. Cyan line -- other ships in the area. Pixels with higher intensity indicate RoIs of the analyzed ship and ships in its neighborhood, if present. }
\label{med_sea_anomal_cont_1}
\end{figure*}

\begin{figure*}
    \centering
    \includegraphics[width=1.0\linewidth]{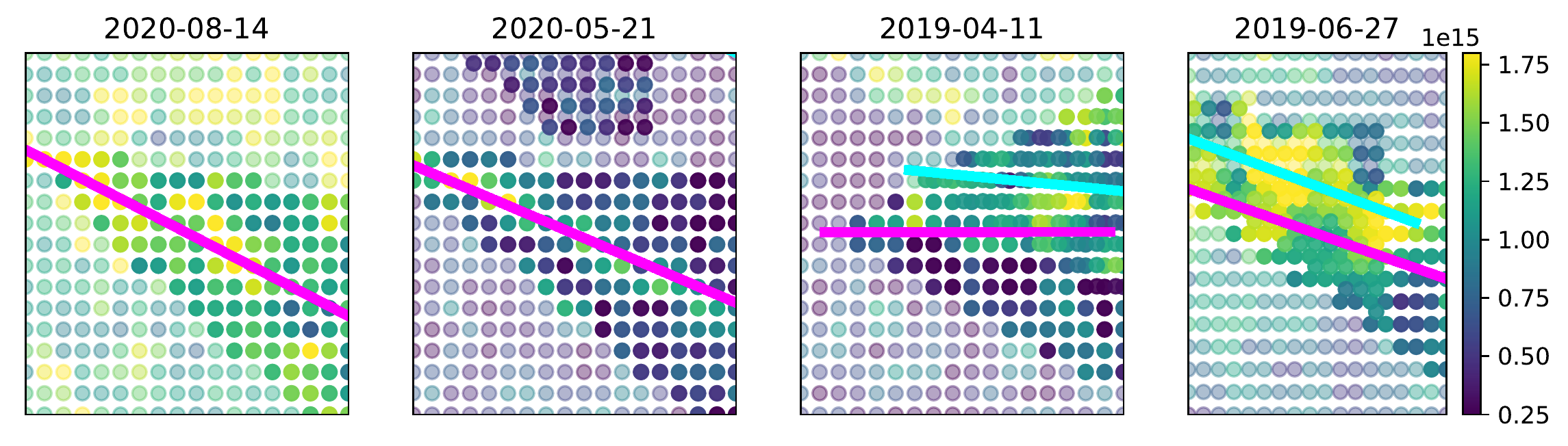}
    \caption{Ship type: Container. Example of an outlying ship 2. Ship length: 363 m. Average ship speed: 17.5 kt. Year of built: 2009\textbackslash 2011 (The information about the year of built of the ship differs in dependence on the used source.). Lines represent shifted ship tracks. Magenta line -- ship of interest. Cyan line -- other ships in the area. Pixels with higher intensity indicate RoIs of the analyzed ship and ships in its neighborhood, if present.}
\label{med_sea_anomal_cont_2}
\end{figure*}

\begin{figure*}
    \centering
    \includegraphics[width=1.0\linewidth]{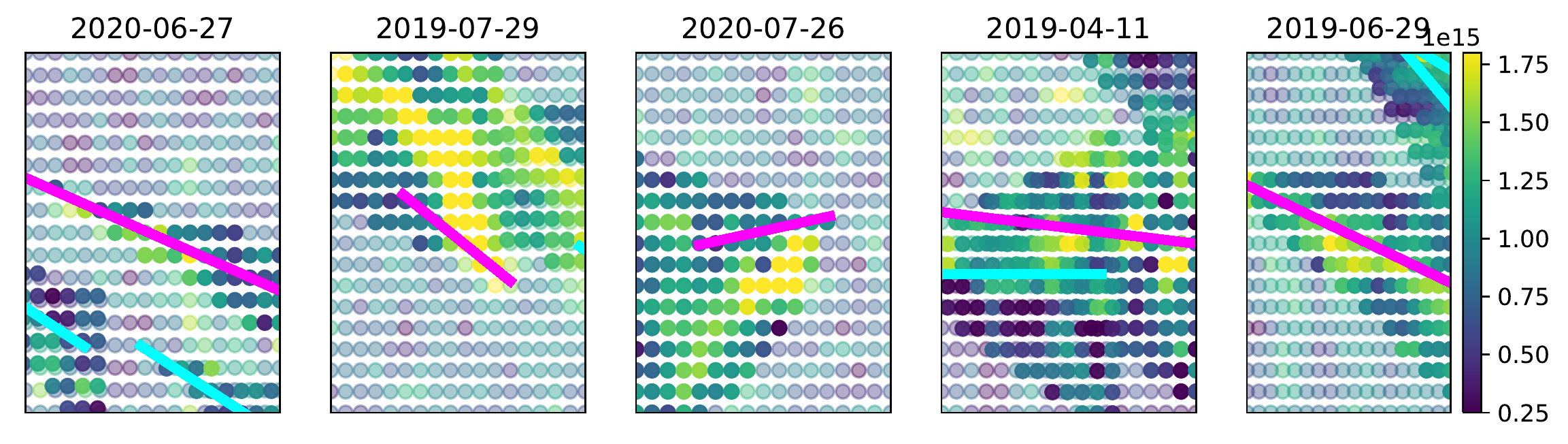}
    \caption{Ship type: Container. Example of an outlying ship 3. Ship length: 397 m. Average ship speed: 18.4 kt. Year of built: 2006. Lines represent shifted ship tracks. Magenta line -- ship of interest. Cyan line -- other ships in the area. Pixels with higher intensity indicate RoIs of the analyzed ship and ships in its neighborhood, if present.}
\label{med_sea_anomal_cont_3}
\end{figure*}

\begin{figure*}
    \centering
    \includegraphics[width=1.0\linewidth]{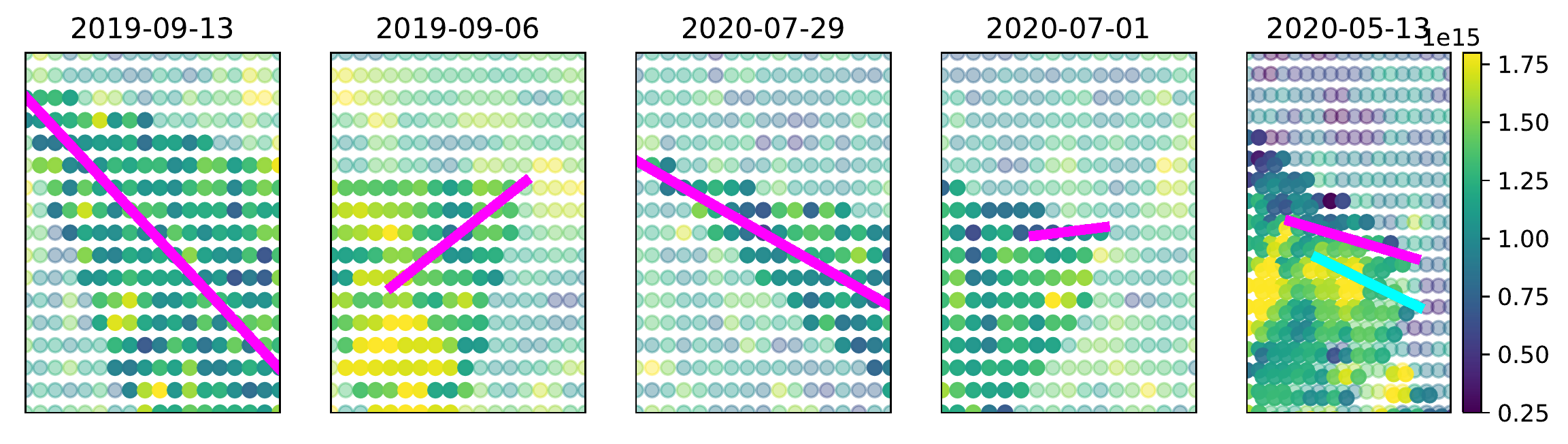}
    \caption{Ship type: Tanker. Example of an outlying ship 1. Ship type: Tanker. Ship length: 180 m. Average ship speed: 15.3 kt. Year of built: 2016. Lines represent shifted ship tracks. Magenta line -- ship of interest. Cyan line -- other ships in the area. Pixels with higher intensity indicate RoIs of the analyzed ship and ships in its neighborhood, if present.}
\label{med_sea_anomal_tanker_1}
\end{figure*}

\begin{figure*}
    \centering
    \includegraphics[width=1.0\linewidth]{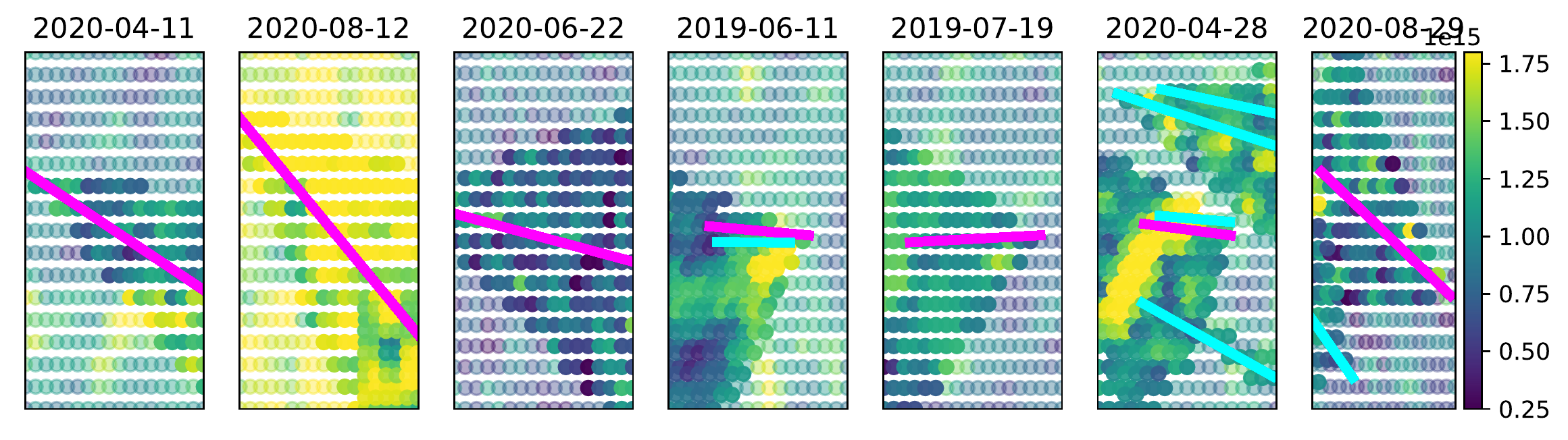}
    \caption{Ship type: Tanker. Example of an outlying ship 2. Ship length: 315 m. Average ship speed: 16.1 kt. Year of built: 2008. Lines represent shifted ship tracks. Magenta line -- ship of interest. Cyan line -- other ships in the area. Pixels with higher intensity indicate RoIs of the analyzed ship and ships in its neighborhood, if present.}
\label{med_sea_anomal_tanker_2}
\end{figure*}

\begin{figure*}
    \centering
    \includegraphics[width=1.0\linewidth]{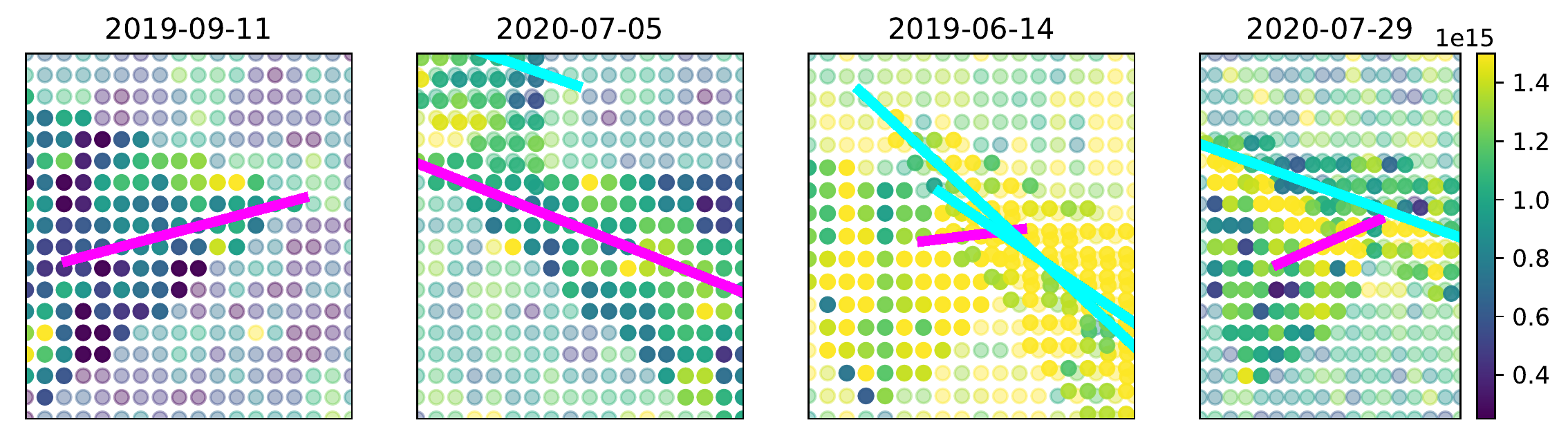}
    \caption{Ship type: Tanker. Example of an outlying ship 3.  Ship length: 179.5 m. Average ship speed: 13 kt. Year of built: 2017. Lines represent shifted ship tracks. Magenta line -- ship of interest. Cyan line -- other ships in the area. Pixels with higher intensity indicate RoIs of the analyzed ship and ships in its neighborhood, if present.}
\label{med_sea_anomal_tanker_3}
\end{figure*}

In order to make final conclusions regarding the ships that were identified by the proposed method as anomalously emitting, as a next step, we visually analyzed the TROPOMI measurements related to those ships.  Figures \ref{med_sea_anomal_cont_1} -- \ref{med_sea_anomal_cont_3} provide the TROPOMI images for red-zone container ships, while Figures \ref{med_sea_anomal_tanker_1} -- \ref{med_sea_anomal_tanker_3} provide the measurements for the red-zone tankers. We also checked the consistency of the results obtained for the red-zone ships (c.f. Table \ref{tab:stds}). 

We can see that for containers, even if there were other ships present in the area, the $\text{NO}_\text{2}$ plumes produced by the ship of interest can always be visually distinguished from the background and from the other plumes. Moreover, for each ship, there are at least two measurement days where there are no other candidates for producing the registered $\text{NO}_\text{2}$ plumes apart from the studied ship. We can also see that the obtained values of the consistency of the results are always from the middle of the data distribution (see Figures \ref{segm_std_hist}, \ref{regr_std_hist}). Therefore, to this point, we do not have reasons to remove any of the selected ships from the list of anomalous emitters.   

In the case of tankers, the situation is different. We can notice that for anomalous emitter with Id 1 the obtained $\sigma(\hat{NO}_2)$ is very high and close to the applied cut-off threshold. Analyzing Figure \ref{med_sea_anomal_tanker_1}, we can see that the cause of such a high standard deviation is a potential plume of land origin present in one of the ship's RoI (measurement from 2019-09-06). This makes us conclude that the given ship should be removed from the list of potential anomalous emitters.

For the tanker with Id 2, both $\sigma(\hat{NO}_2)$ and $\sigma(diff)$ are within the distributions. However, analyzing Figure \ref{med_sea_anomal_tanker_2}, we can see that at least three times the ship was observed at the moment of crossing the plume from another ship. From this, we conclude that the ship was attributed to the list of anomalous emitters due to the consistent $\text{NO}_\text{2}$ contributions of nearby ships, rather than consistent production of anomalously high levels of emission.  

Finally, for the tanker with Id 3, we observe that one of the images (measurement date: 2019-06-14) has a
relatively high level of background $\text{NO}_\text{2}$ concentration. As a consequence, we cannot take it into consideration in our analysis. The rest of the images, nevertheless, show visually distinguishable $\text{NO}_\text{2}$ plume that can be attributed to the ship of our interest. 
Consequently, we conclude that we do not have reasons to remove a given ship from the list of anomalous emitters.

\begin{table*}[]
    \centering
    \begin{tabular}{cccc}
    \toprule
        \textbf{Ship type} & \textbf{Variable} & \textbf{Strong emitters} & \textbf{Weak Emitters}  \\
        \midrule
         Tanker & Year of built & 2013 $\pm$ 5 & 2009 $\pm$ 4 \\
         
        & Ship length [m] & 224 $\pm$  78 & 253 $\pm$ 66  \\
        
        & Ship speed [kt] & 14.8 $\pm$ 1.5 & 14.8 $\pm$ 1.6 \\
        
        & Wind speed [m/s] & 4.9 $\pm$  0.4 & 5.0 $\pm$ 0.7 \\
        
        & Average IoU  & 0.07 $\pm$ 0.1 & 0.05 $\pm$ 0.06 \\
        \hline
        Container & Year of built &  2008 $\pm$ 2 & 2012 $\pm$ 5  \\
        
        & Ship length [m] & 386 $\pm$ 20 & 340 $\pm$ 70 \\
        
        & Ship speed [kt] & 18.5 $\pm$ 1. & 17.1 $\pm$ 1.7 \\
        
        & Wind speed [m/s] & 4.8 $\pm$ 0.5 & 5.1 $\pm$ 0.8   \\
        
        & Average IoU  & 0.07 $\pm$ 0.02 & 0.04 $\pm$ 0.04 \\
    \bottomrule   
    \end{tabular}
    \caption{Statistical summary for important factors that influence levels of produced $\text{NO}_\text{2}$ for ships that by both models were identified as strong and weak emitters. IoU stands for Intersection over Union.}
    \label{tab:averages}
\end{table*}

\begin{table}[]
    \centering
    \begin{tabular}{cccc}
    \toprule
        \textbf{Ship type}  & \textbf{Variable}& \textbf{F statistic} & \textbf{p-value}  \\
         \midrule
         Tanker & Year of built & 2.3 & 0.13 \\
         &Ship length & 0.48 & 0.49 \\
         &Ship speed & 0.004 & 0.95 \\
         &Wind Speed & 0.12 & 0.72 \\
         &Average IoU & 0.4 & 0.53 \\
         \hline
         Container & Year of built & 1.7 & 0.19 \\
         & Ship length & 0.24 & 0.27 \\
         & Ship speed & 1.95 & 0.16 \\
         & Wind Speed & 0.53 & 0.47 \\
         & Average IoU & 1.32 & 0.25 \\
    \bottomrule   
    \end{tabular}
    \caption{One way ANOVA for the significance of the statistical difference between samples of ships identified as strong and weak emitters. IoU stands for Intersection over Union.}
    \label{tab:ANOVA}
\end{table}

\subsection{Decision bias}

To select the anomalously emitting ships, we combined the results of two independently trained models: a regression model for ship $\text{NO}_\text{2}$ estimation and a model of ship plume segmentation. 
Taking this into account, as a final step of the analysis, we would like to know if such a model fusion did not create any decision bias that would pre-determine the attribution of a certain ship to a class of strong or weak emitters. For this, we decided to study five variables that are interesting from the point of view of result interpretability. Three of the selected variables (ship length, ship speed, and wind speed) were features of both regression and segmentation models. Another two variables (Year of built -- stands for the ship built year, and Average IoU -- stands for an average score of Intersection over Union of ships RoI with the RoI of other ships\footnote{Given two areas of interest, IoU is computed as the surface of their overlap divided by the surface of their joint area.}) were not a part of any model\footnote{The variables were tested in the preliminary phase of our regression model experiments but were removed due to the negative impact on model performance.} but can have a potential influence on the attribution of a ship to a class of weak or strong emitters.

To check the potential presence of decision bias, for each studied ship type, we compared the averages of the above-mentioned features (see Table \ref{tab:averages}) and performed a univariate one-way ANOVA test (Table \ref{tab:ANOVA}), analyzing the statistical significance of the differences between the values of the variables from two groups of ships -- strong or weak emitters. From the obtained results, we conclude that none of the analyzed variables had a statistically significant influence on attributing a certain ship to a class of strong or weak emitters. This implies the absence of decision bias related to these variables. 

\section{Discussion}\label{discussion}

In this study, we presented a method for detecting anomalously $\text{NO}_\text{2}$ emitting ships. For this, we first introduced a specifically designed regression model for the prediction of the level of $\text{NO}_\text{2}$ of an individual ship. 
The presented regression model estimates $\text{NO}_\text{2}$ levels using remote sensing --- it does not require manual labeling and is validated using TROPOMI data directly rather than by using emission proxy of limited flexibility like it has been done before. 
The model was trained on historical TROPOMI data. Therefore, the developed model is able to predict how much  $\text{NO}_\text{2}$ is expected to be registered by the TROPOMI sensor for a ship with certain properties operating at certain atmospheric conditions.
Ships for which the real registered amount of $\text{NO}_\text{2}$ is higher than predicted by the model were of our highest interest.

However, there are many factors that potentially can result in a situation when the TROPOMI sensor registers more $\text{NO}_\text{2}$ than could have been expected: atmospheric chemistry processes, intersection with another plume, $\text{NO}_\text{2}$ plumes produced by land-based sources, etc. To minimize the effect of such factors, we studied the ship performance over an extended period of time (2 years), creating ships' profiles composed of multiple observations. We checked how consistent the obtained results are -- if the results' variability is too high, such a profile cannot be used for making conclusions. 

We bear in mind that due to the many factors (i.a. atmospheric processes affecting  $\text{NO}_\text{x} \rightarrow \text{NO}_\text{2}$ transformations, and other model-based priors that are part of VCD$_{trop}$ calculations, as described in \ref{tropomi_sect}) the TROPOMI $\text{NO}_\text{2}$ columns cannot be treated as a ground truth that precisely quantifies the amount of $\text{NO}_\text{x}$ emitted by a studied ship. Therefore, to use the proposed regression model for the detection of anomalously emitting ships, we combined it with another machine learning model for ship  $\text{NO}_\text{2}$ estimation. This is a ship plume segmentation model that takes into account the spatio-temporal distribution of a ship plume, which compensates for the main disadvantage of the proposed regression model, where the $\text{NO}_\text{2}$ within the ship sector is summarized rather than treated pixel-wise. The used plume segmentation model is trained with human-made annotations and is validated using a theoretical ship emission proxy. All above-mentioned allow us to assume that the results produced with the segmentation model will be independent of those obtained with the regression model applied to the same dataset. The ships that were nominated as repeatedly highly emitting by both of the models, we considered potentially anomalously emitting.

For ships identified as anomalous emitters, we visually inspected the corresponding TROPOMI measurements. For 4 out of 6 ships that were initially classified as anomalous emitters we did not find any major factors unrelated to that ship that could have affected the obtained results. The two other ships, however, were selected as anomalous emitters as a result of repeated intersection with the plumes from another source or land outflow. Even though measures to prevent such situations were taken upon the construction of the method, due to the high irregularities of both atmospheric chemistry processes and ship trajectories, it is still very difficult to fully eliminate from the analysis or properly process such signal interference cases. 

As a last step of the analysis, we checked whether the performed fusion of models did not create the additional biases that predetermine the attribution of a ship to a class of anomalous emitters. For this, we performed a one-way ANOVA test of group differences. We selected a set of variables that were model features of both used machine learning approaches and a set of variables that were not used by any of the models but can potentially introduce the bias into results. None of the studied variables showed a statistically significant difference between the group of high and low emitters. 
Nevertheless, the used one-way ANOVA test allows us to study the influence of single variables only, ignoring the inter-group effects that may come from the combination of variables. The small number of anomalous ships restricts us from the application of multivariate approaches to this problem.

In this study, by observing a ship throughout several days of performance, we took into account the temporal consistency of the levels of emission from that ship. As a future work, we propose to examine also spatial consistency. This can be done by performing similar experiments on multiple parts of a shipping route. However, due to the different meteorological conditions, different levels of concentration of outflows from land, as well as possibly different priors (such as AMF) taken for VCD calculations, such an extension will require the training of separate regression and segmentation machine learning models, which in case of segmentation model comes with labeling a new dataset. 

Finally, the dynamic of the atmospheric processes affects how fast and how much $\text{NO}_\text{2}$ will be created from $\text{NO}_\text{x}$ emitted by a ship. In this study, we implicitly addressed the atmospheric chemistry processes by using features such as the month the observation took place (seasonability) and solar angle. As an improvement of the current methodology, we suggest the explicit modeling of atmospheric chemistry through a regression model. For instance, by the introduction of such features as ozone concentration or air temperature. 

\section{Conclusions}\label{conclusions}
In this study, we applied a combination of machine learning-based methods on TROPOMI satellite data and presented an approach for automatic identification of anomalously  $\text{NO}_\text{2}$ emitting ships. 
Our approach allows the automatic processing of a huge amount of satellite remote sensing data in order to select ships for inspection the ships that consistently emit more than can be inferred based on their properties and sailing conditions. With the proposed methodology, the selected cases for inspection are based on multi-day observations of a given ship. With this, we harvest the main advantage of satellite observations over the existing approaches for ship compliance monitoring, with which the decisions have to be made on the basis of a single observation only. The proposed methodology can be used as a recommendation system for ship inspectors.  

\begin{acks}
This work is funded by the Netherlands Human Environment and Transport Inspectorate, the Dutch Ministry of Infrastructure and Water Management, and the SCIPPER project, which receives funding from the European Union’s Horizon 2020 research and innovation program under grant agreement Nr.814893.
\end{acks}

\typeout{}
\bibliographystyle{ACM-Reference-Format}
\bibliography{literature}


\appendix
\begin{appendices}
\section{Regression model}\label{Aregr_dataset}
\begin{figure*}
    \centering
    \includegraphics[width=1.0\linewidth]{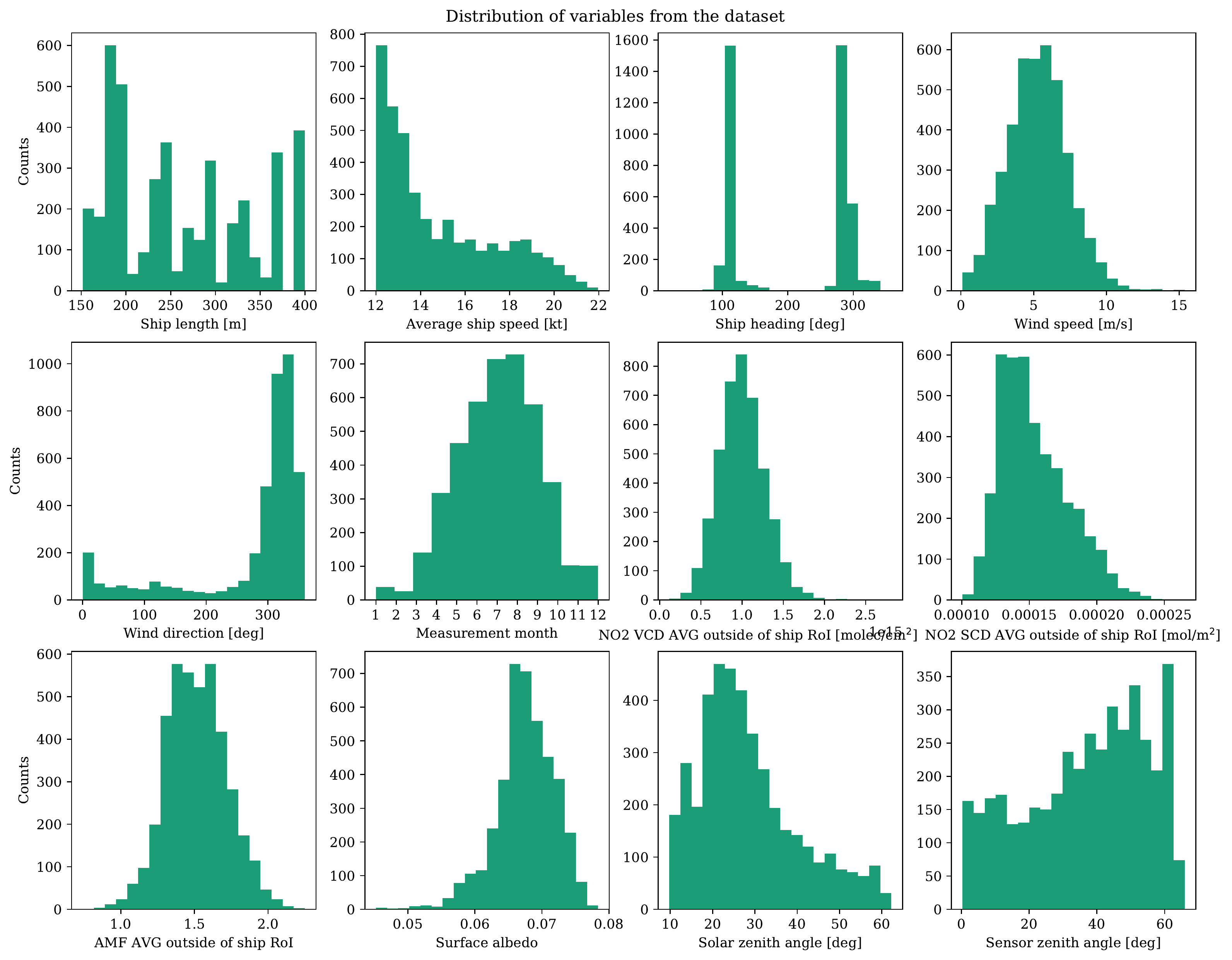}
    \caption{Distribution of variables of the dataset used in this study.}
\label{var_hists}
\end{figure*}

The dataset used for the regression model is composed of 4153 rows (aggregated ship plume images). Figure \ref{var_hists} presents the histograms of the distributions of the variables from the regression model dataset.

\section{segmentation model dataset}\label{Asegm_dataset}
The dataset used for the training of ship plume segmentation model is composed of 68 days of $\text{NO}_\text{2}$  TROPOMI measurements taken between 1 April 2019 and 31 December 2019. The dataset covers the same area in the Mediterranean Sea as the regression model dataset (see Section \ref{tropomi_sect}).

\section{Segmentation model optimization}\label{Asegm_model}
In \cite{rs14225809} it was shown that the highest performance quality of the ship plume segmentation task was achieved with XGBoost classifier. Therefore, in this study, for the task of ship plume segmentation, we use XGBoost model and optimize it using the methodology from the original article. The hyperparameters  The obtained cross-validation-averaged average precision score is equal to 0.753. For the extensive reports of the model performance evaluation, we direct the reader to \cite{rs14225809}.

\end{appendices}

\end{document}